\documentclass[10pt,twocolumn,letterpaper]{article}

\usepackage[pagenumbers]{cvpr} 
\usepackage{url}            
\usepackage{booktabs}       
\usepackage{amsfonts}       
\usepackage{nicefrac}       
\usepackage{microtype}      
\usepackage{xcolor}         
\usepackage{wrapfig} 

\usepackage{adjustbox}
\usepackage{diagbox}
\usepackage{multirow}
\usepackage{makecell}

\usepackage{times}
\usepackage{epsfig}
\usepackage{graphicx}
\usepackage{amsmath}
\usepackage{amssymb}
\usepackage{colortbl}
\usepackage{xspace}

\usepackage{ragged2e}
\usepackage{float}
\usepackage{bbding}
\usepackage{amssymb}  
\usepackage[accsupp]{axessibility}
\usepackage{bbding}

\usepackage{hyperref}
\usepackage{gensymb}

\title{\vspace{-1.5cm}ManualVLA: A Unified VLA Model for Chain-of-Thought \\Manual Generation and Robotic Manipulation}

\author{
\hspace{-4mm}\textbf{Chenyang Gu\thanks{Equal contribution $^{\dagger}$Project leader \textsuperscript{\Envelope}Corresponding author.}} \hspace{0.1mm} \textsuperscript{\rm 1}, \hspace{0.1mm}
\textbf{Jiaming Liu$^{*\dagger}$\textsuperscript{\rm 1}}, \hspace{0.1mm}
\textbf{Hao Chen$^{*\dagger}$\textsuperscript{\rm 2}}, \hspace{0.1mm}
\textbf{Runzhong Huang$^*$\textsuperscript{\rm 1}}, \hspace{0.1mm}
\vspace{0.05cm} 
\textbf{Qingpo Wuwu}\textsuperscript{\rm 1},  \hspace{0.1mm}
\textbf{Zhuoyang Liu\textsuperscript{\rm 1}}, \hspace{0.1mm}\\
\textbf{Xiaoqi Li}\textsuperscript{\rm 1},  \hspace{0.1mm}
\textbf{Ying Li}\textsuperscript{\rm 1}, \hspace{0.1mm}
\textbf{Renrui Zhang\textsuperscript{\rm 2}}, \hspace{0.1mm}
\textbf{Peng Jia\textsuperscript{\rm 3}}, \hspace{0.1mm}
\textbf{Pheng-Ann Heng\textsuperscript{\rm 2}},  \hspace{0.1mm}
\textbf{Shanghang Zhang\textsuperscript{\Envelope}~\textsuperscript{\rm 1}} 
\vspace{0.2cm} \\
\textsuperscript{\rm 1}State Key Laboratory of Multimedia Information Processing,
School of Computer Science, \\Peking University  
\textsuperscript{\rm 2}The Chinese University of Hong Kong \textsuperscript{\rm 3}Simplexity Robotics \\
\textbf{Project web page:} \href{https://sites.google.com/view/maunalvla/}{https://sites.google.com/view/maunalvla}.
}

\begin{document}
\maketitle
\begin{abstract}

Vision–Language–Action (VLA) models have recently emerged, demonstrating strong generalization in robotic scene understanding and manipulation. 
However, when confronted with long-horizon tasks that require defined goal states, such as LEGO assembly or object rearrangement, existing VLA models still face challenges in coordinating high-level planning with precise manipulation.
Therefore, we aim to endow a VLA model with the capability to infer the “how” process from the “what” outcomes, transforming goal states into executable procedures.
In this paper, we introduce ManualVLA, a unified VLA framework built upon a Mixture-of-Transformers (MoT) architecture, enabling coherent collaboration between multimodal manual generation and action execution.
Unlike prior VLA models that directly map sensory inputs to actions, we first equip ManualVLA with a planning expert that generates intermediate manuals consisting of images, position prompts, and textual instructions. Building upon these multimodal manuals, we design a Manual Chain-of-Thought (ManualCoT) reasoning process that feeds them into the action expert, where each manual step provides explicit control conditions, while its latent representation offers implicit guidance for accurate manipulation.
To alleviate the burden of data collection, we develop a high-fidelity digital-twin toolkit based on 3D Gaussian Splatting, which automatically generates manual data for planning expert training.
ManualVLA demonstrates strong real-world performance, achieving an average success rate 32\% higher than the previous hierarchical SOTA baseline on LEGO assembly and object rearrangement tasks.

\end{abstract}    
\vspace{-0.3cm}
\section{Introduction}
\label{sec:intro}

\begin{figure}[t]
\includegraphics[width=0.46\textwidth]{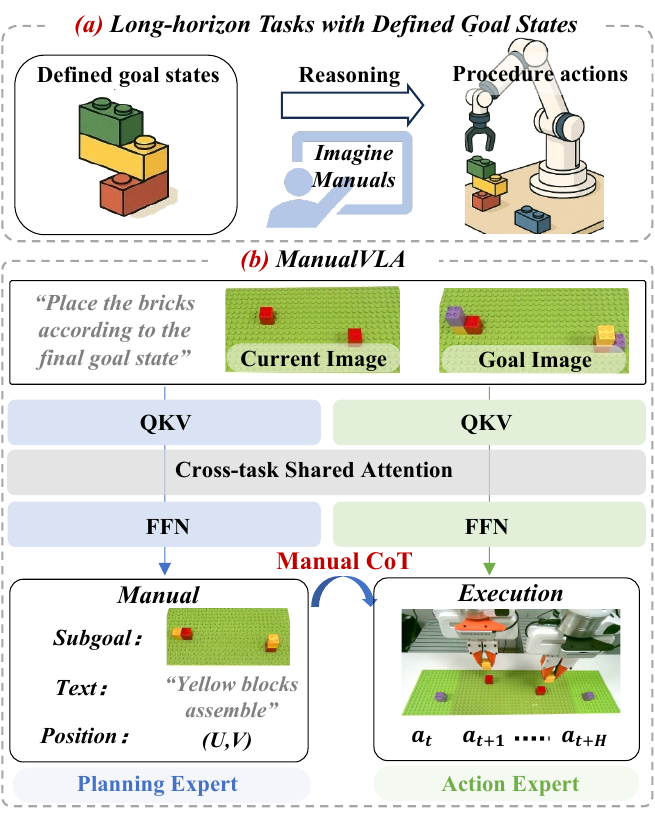}
\centering
\vspace{-0.2cm}
\caption{\textbf{Overview.}
\textcolor{red}{(a)} Long-horizon tasks with predefined goal states, such as LEGO assembly or object rearrangement, pose a significant challenge for intelligent robots, as they require not only imagining procedural manuals but also executing precise manipulations based on them.
\textcolor{red}{(b)} We address such tasks by introducing ManualVLA, a unified VLA model built upon a MoT architecture, which enables coherent collaboration between multimodal manual and action generation via a designed Manual Chain-of-Thought.
}
\label{fig:intro}
\vspace{-0.3cm}
\end{figure}

Recently, building on internet-scale pretrained vision-language models~\cite{alayrac2022flamingo, karamcheti2024prismatic}, vision-language-action (VLA) models have emerged~\cite{rt22023arxiv, kim2024openvla} and are trained on robot demonstrations to predict control actions.
These models exhibit impressive capabilities in robotic scene understanding~\cite{yang2025instructvla, lin2025onetwovla, liu2024robomamba} while demonstrating strong generalization and manipulation performance in out-of-lab scenarios~\cite{intelligence2025pi05visionlanguageactionmodelopenworld, liu2025hybridvla, bjorck2025gr00t}.
Such advances have significantly accelerated progress toward developing generalist robotic agents.
However, when confronted with long-horizon tasks that require precisely defined goal states (Figure~\ref{fig:intro} {\textcolor{red}{(a)}}), such as LEGO assembly or object rearrangement, VLA models remain highly challenged.
The difficulty arises from two aspects: (1) the VLA model must execute precise manipulation strictly aligned with the predefined final scene or object configuration, and (2) it must integrate long-horizon planning with fine-grained control, while maintaining generalization to diverse real-world environments.

In contrast, for humans, upon accumulating sufficient manipulation experience, they can perform long-horizon, goal-specific tasks independently, without relying on further demonstrations.
For example, humans can assemble LEGO structures by inferring the target configuration, or reorganize household objects according to a predefined spatial arrangement.
This ability stems from humans’ intuitive inference of intermediate cues, reasoning over spatial and causal relations, and decomposition of tasks into coherent, subgoal-directed manipulation steps.
Recently, some hierarchical methods have attempted to emulate this ability by relying on detailed manuals~\cite{long2025checkmanual} or human demonstration videos~\cite{papagiannis2025r+, jain2024vid2robot, bahl2022human, wang2024dexcap}. However, such approaches are often limited in generalization to unseen final goal states and increase the dependence on human involvement.
This naturally leads to a question: \textit{``Is it possible to endow a VLA model with the capability to infer the procedural “how” from the desired “what,” thereby transforming a predefined final goal into a sequence of coherent and precise execution steps?"}

To this end, we propose \textbf{ManualVLA}, a unified VLA model based on a Mixture-of-Transformers (MoT)~\cite{liang2024mixture} architecture, capable of generating multimodal manuals and actions directly from a final goal state.
Since only the final state is provided and intermediate steps are unknown, previous VLA models that map sensory inputs directly to actions struggle with such long-horizon tasks. In contrast, as shown in Figure~\ref{fig:intro} \textcolor{red}{(b)}, ManualVLA selectively activates planning and action experts within a unified framework for subgoal manual and action generation.
Specifically, ManualVLA is equipped with a planning expert to generate intermediate manuals that integrate images, position prompts, and textual instructions. 
These manuals are then used in our proposed \textbf{Manual Chain-of-Thought (ManualCoT)} reasoning strategy, which guides the action expert by treating each subgoal step as an explicit condition for precise execution.
Furthermore, a cross-task shared attention mechanism between the two experts enables long-context interactions between manual-generation features and action generation, providing implicit guidance for coherent manipulation.

\textbf{To train ManualVLA}, we first leverage the generative capability of VLM model (e.g., Janus-Pro~\cite{wu2024janus}) and fine-tune it on self-collected and simulation-synthesized manual data to acquire the planning expert capability.
Due to the substantial uncertainty in the final goal states, a large amount of data is required to train the model with sufficient world knowledge for effective task planning. To alleviate the burden of data collection, we develop a high-fidelity digital-twin toolkit based on 3D Gaussian Splatting~\cite{2023_8_08-3dgs_for_real_time_radiance_field_rendering}, which automatically generates manual data for planning expert training.
Meanwhile, we pretrain the action expert on large-scale open-source robotic datasets comprising over 400K trajectories~\cite{open_x_embodiment_rt_x_2023, khazatsky2024droid}. Benefiting from the rich manual conditions incorporated during action generation, ManualVLA requires only around 100 trajectories to achieve generalizable manipulation while finetuning on downstream tasks. 
Experimental results demonstrate that, when confronted with long-horizon and complex LEGO assembly and object rearrangement tasks, 
ManualVLA not only generates accurate manuals but also achieves an average manipulation success rate of 32\% higher than existing hierarchical SOTA baseline.
Note that ManualVLA also achieves state-of-the-art (SOTA) performance on other general manipulation tasks.
In summary, our contributions are as follows:

\begin{itemize}

    \item We address long-horizon tasks with precisely defined goal states by introducing ManualVLA, a unified VLA model built upon a MoT architecture that supports coherent multimodal manual generation and action execution.

    \item We design a Manual Chain-of-Thought (ManualCoT) reasoning process that translates generated manuals into precise actions, where each manual step provides explicit control conditions and its latent representation offers implicit guidance for manipulation.

    \item Equipped with the proposed training strategy, ManualVLA demonstrates strong real-world performance, achieving superior manipulation accuracy and generalization compared with previous SOTA baselines on downstream tasks.
\end{itemize}

\section{Related Work}

\noindent\textbf{Vision-language models (VLMs)}~\cite{alayrac2022flamingo,radford2021learning,karamcheti2024prismatic} have achieved strong multimodal reasoning by learning from internet-scale image-text data. Building on this progress, VLA models~\cite{kim2024openvla,rt22023arxiv} have emerged as a promising approach for robot learning, enabling end-to-end mapping from multimodal observations to control signals.
Subsequent works have further advanced VLA models by incorporating richer sensory understanding~\cite{qu2025spatialvla, li2025pointvlainjecting3dworld, liu2025mla, li20253ds}, exploring more robust action generation~\cite{black2024pi_0, intelligence2025pi05visionlanguageactionmodelopenworld, wen2024diffusion, li2024cogact, liu2025hybridvla}, and developing optimized inference strategies~\cite{pertsch2025fast, wen2025tinyvla, liu2024robomamba} as well as dual-system paradigms~\cite{zhang2024hirt, figure2024helix, bjorck2025gr00t, chen2025fast}.
Notably, recent approaches like MoTVLA~\cite{huang2025motvlavisionlanguageactionmodelunified} and F1-VLA~\cite{lv2025f1} have introduced MoT architectures into general robotic manipulation.
However, when faced with long-horizon tasks that require precise goal specifications, these models continue to struggle.
To reconcile this issue, several works~\cite{hu2024video,zhou2024robodreamer,ye2024latent,liang2024dreamitate} have incorporated visual world modeling into VLA architectures to enable reasoning about the future images.
They typically decouple long-horizon progress into intermediate steps either by generating explicit pixel-level subgoals~\cite{google2024pivot,wu2023-GR1} or by formulating compressed token representations~\cite{zhang2025dreamvla,bu2025agibot,gao2025adaworld,li2025unified}.
Nevertheless, these models struggle to capture the relationship between subgoals and fine-grained control. In contrast, we introduce a comprehensive Chain-of-Thought (CoT) reasoning process that combines both explicit and implicit cues to transform the generated manuals into precise actions.

\noindent\textbf{Final goal conditioned manipulation}, where robots must reach specified target states from given initial configurations, such as LEGO assembly or object rearrangement, represents a challenge in embodied AI.
A prominent research direction uses human hand videos~\cite{bahl2022human,shaw2023videodex,jain2024vid2robot} to present the desired intermediate procedure.
Methods such as Vid2Robot~\cite{jain2024vid2robot} and DexCap~\cite{wang2024dexcap} extract manipulation trajectories from egocentric videos, transferring human dexterity to robotic control.
Meanwhile, several works~\cite{wang2022translating,pun2025generating,zhang2025manual} exploit operation manuals or goal-state descriptions as guidance.
CheckManual~\cite{long2025checkmanual} conditions robot policies on predefined instruction manuals, while ~\cite{wu2022targf, zeng2024lvdiffusor} utilize the final target scene configuration to inform execution goal.
However, providing hand videos, human-crafted manuals, or additional reasoning models introduces extra human effort and computational cost, limiting the practicality of these approaches.
In contrast, we make the first attempt to address long-horizon, goal-conditioned manipulation through a unified VLA model that enables coherent collaboration between multimodal manual generation and action execution.

\vspace{-0.1cm}
\section{ManualVLA}
\label{sec:manualvla}
In this section, we first introduce the fundamentals of Vision-Language-Action (VLA) models in Section \ref{sec:4.1}. Then, Section \ref{sec:4.2} presents the architectural details of ManualVLA, followed by Section \ref{sec:4.3}, which elaborates on its working principles. Sections~\ref{sec:4.4} and~\ref{sec:3.5} describe the training strategies of ManualVLA and the digital-twin toolkit, respectively.

\subsection{Preliminary}
\label{sec:4.1}
\begin{figure*}[t] 
    \centering
    \includegraphics[width=\textwidth]{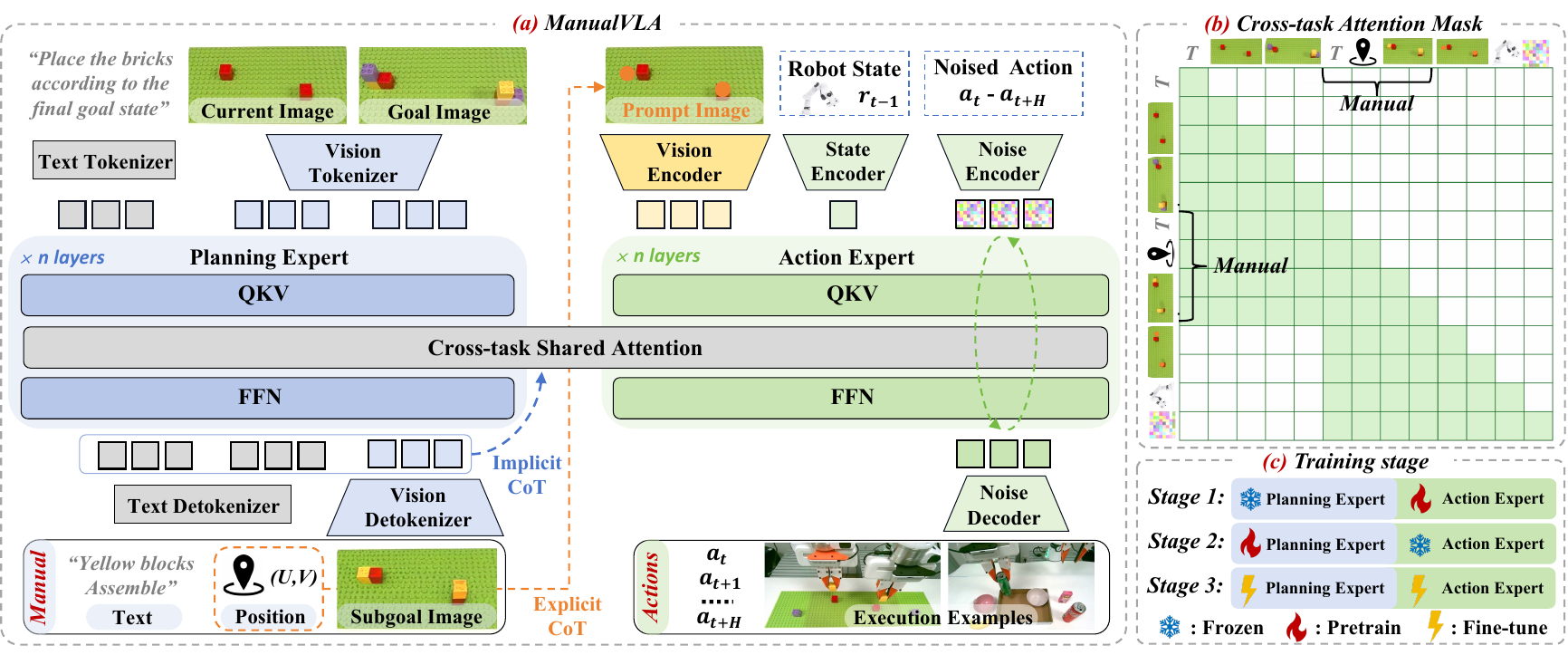} 
    \vspace{-0.7cm}
    \caption{\textbf{Framework of ManualVLA.}
\textcolor{red}{(a)} To accomplish long-horizon tasks with defined goal states, we propose ManualVLA, a unified VLA model built upon a MoT architecture. The framework consists of two experts: a planning expert responsible for generating multimodal manuals, and an action expert responsible for predicting precise actions.
The planning expert processes human instructions, the current image, and the final goal image to generate intermediate manuals that combine next-step image, positions, and sub-task instructions. We introduce an explicit CoT reasoning process, where each positional indicator serves as a visual prompt embedded into the observation of the action expert.
\textcolor{red}{(b)} Along with the cross-task shared attention mechanism and the designed attention mask, the generated manual tokens are also used as conditioning signals for action generation, enabling an implicit CoT reasoning process that effectively guides the action expert.
\textcolor{red}{(c)} ManualVLA adopts a three-stage training strategy that aligns the planning and action experts for effective collaboration.
    }
    \label{fig:method}
    \vspace{-0.25cm}
\end{figure*}

VLA models integrate visual, linguistic, and proprioceptive inputs to generate robot control signals, exhibiting strong generalization in diverse manipulation tasks~\cite{kim2024openvla,black2024pi_0}.
Despite their impressive abilities, existing VLA models often struggle with long-horizon tasks with defined goal states, reflecting their limited world knowledge for planning the intermediate progress required to achieve such goals.
To address this limitation, we aim to endow the VLA model with a human-like capability to infer the procedural “how” from the desired “what”. Therefore, we propose ManualVLA $\pi_\theta$, a unified VLA model equipped with world knowledge that first reasons about multimodal manuals describing the task procedure, and subsequently generates the corresponding actions for execution.
As shown in Figure.~\ref{fig:method}, given the language instruction $l$ together with 
the images of the current state $\mathcal{I}_t^{\text{current}}$ and the final goal state $\mathcal{I}^{\text{goal}}$, 
ManualVLA first generates a manual that consists of the textual description of the target objects $\hat{l}_t$, 
their target 2D coordinates $p_t$, and the corresponding subgoal image $\mathcal{I}_t^{\text{subgoal}}$:
\begin{equation}
\pi_{\theta}(\mathcal{I}_t^{\text{subgoal}}, p_t, \hat{l}_t \mid \mathcal{I}^{\text{goal}}, \mathcal{I}_t^{\text{current}}, l).
\end{equation}

Based on this manual, we construct a prompted image $\mathcal{I}_t^{\text{prompt}}$ 
by overlaying the target object’s final position as a mask on the current scene image $\mathcal{I}_t^{\text{current}}$. 
Finally, the model takes the robot state $s_t$ and the prompted image $\mathcal{I}_t^{\text{prompt}}$, together with $\mathcal{F}_{t}^{\text{subgoal}}$, $\mathcal{F}^{p}_{t}$, and $\mathcal{F}^{\hat{l}}_{t}$ (the key and value features 
stored during the generation of $\hat{l}_t$, $p_t$, and $\mathcal{I}_t^{\text{subgoal}}$), as conditional inputs for modeling the action chunk $a_{t:t+h}$, enabling explicit subgoal-guided action generation.

\begin{equation}
\pi_{\theta}(a_{t:t+h} \mid s_t, \mathcal{I}_t^{\text{prompt}}, \mathcal{F}_{t}^{\text{subgoal}}, 
\mathcal{F}^{p}_{t}, \mathcal{F}^{\hat{l}}_{t}).
\end{equation}

\subsection{Model Architecture}
\label{sec:4.2}
ManualVLA adopts Janus-Pro \cite{chen2025janus} as its foundation model due to its strong capability in general multimodal understanding and generation. 
Unlike atomic or short-horizon manipulation tasks, our goal is to perform long-horizon, goal-specific tasks, where the model must not only generate detailed manuals but also predict precise actions and enable effective interaction between the two processes.
To achieve this, we extend the basic VLM into a Mixture-of-Transformers (MoT) architecture, forming a unified VLA model that integrates collaborative planning and action experts.
We next detail the key components of our ManualVLA.

\textbf{Vision Tokenizer and Encoder.} 
As shown in Figure~\ref{fig:method}, given the distinct characteristics of discrete and continuous image-injection paradigms, as well as the differing demands of manual and action generation, ManualVLA employs two separate visual modules: a VQ-based vision tokenizer for manual generation and a continuous vision encoder for action generation.
For the vision tokenizer, ManualVLA adopts an encoder-quantizer-decoder architecture following VQGAN~\cite{esser2021taming}. The encoder and decoder are convolutional networks with a downsampling factor of 16, and the quantizer maintains a codebook $\mathbf{Z} \in \mathbb{R}^{16{,}384 \times 8}$.
For the vision encoder, ManualVLA uses SigLIP-Large~\cite{zhai2023sigmoid} with an input resolution of 384 to extract high-dimensional semantic features from input images.

\textbf{Mixture-of-Transformers LLM.} The base language model in ManualVLA is DeepSeek-LLM 1.5B~\cite{bi2024deepseek}.
To integrate the distinct capabilities of manual generation and action generation, we construct a MoT architecture atop this LLM. 
The proposed MoT extends the standard Transformer by introducing task-specific parameter sets for all non-embedding components, including feed-forward networks (FFN), attention projections, and layer normalizations, yielding the planning and action experts illustrated in Figure~\ref{fig:method} \textcolor{red}{(a)}.
When faced with complex and long-horizon tasks, this design enables the VLA model to efficiently handle heterogeneous tasks while preserving its ability to learn cross-task dependencies within a unified framework.

To formally describe this mechanism, we take a single MoT layer as an example.
Let \(x=(x_1,\dots,x_n)\) denote the input token sequence, where each token \(x_i\) is assigned to exactly one task category \(t_i \in \mathcal{T}=\{\text{manual},\text{action}\}\).
For each task, we define task-dependent operator bundles
\(\Theta^t=\{\theta_{\text{attn}}^{t},\theta_{\text{ffn}}^{t}\}\)
and corresponding mappings
\(\Phi_{\text{attn}}^{t}, \Phi_{\text{ffn}}^{t} : \mathbb{R}^d \to \mathbb{R}^d\),
which are applied token-wise depending on the associated task.
Then, a MoT layer acting on a mixed-task sequence can be compactly expressed as:
\begin{equation}\label{eq:mot-layer-abstract}
\mathrm{MoT}_{\Theta}(x)
\;=\;
x
\;+\;
\mathcal{N}_{\text{ffn}}^{t(\cdot)}
\!\Big(
\Phi_{\text{ffn}}^{t(\cdot)}
\big(
x
\;+\;
\mathcal{N}_{\text{attn}}^{t(\cdot)}\big(\Phi_{\text{attn}}(x)\big)
\big)
\Big),
\end{equation}
where the notation \(t(\cdot)\) indicates that each token at position \(i\) uses its corresponding task parameters and \(\mathcal{N}_{\text{attn}}^{t}\), \(\mathcal{N}_{\text{ffn}}^{t}\) denote task-specific layer-normalization operators.

Finally, to define the global attention operator~\cite{deng2025emerging}, let \(X \in \mathbb{R}^{n\times d}\) denote the matrix form of the input sequence \(x=(x_1,\dots,x_n)\), where each row corresponds to a token embedding. The operator is then given by
\begin{equation}\label{eq:global-attn-abstract}
\begin{aligned}
Q &= X W_Q^{t(\cdot)},\quad
K = X W_K^{t(\cdot)},\quad
V = X W_V^{t(\cdot)},\\[4pt]
A &= \mathrm{softmax}\!\Big(\frac{QK^{\top}}{\sqrt{d_k}}\Big), \quad\Phi_{\text{attn}}(x) = (A V)\, W_O^{t(\cdot)},
\end{aligned}
\end{equation}
where the projections \(W_Q^{t(\cdot)},W_K^{t(\cdot)},W_V^{t(\cdot)},W_O^{t(\cdot)}\)
are selected per token according to its task \(t_i\),
while the attention weight matrix \(A\) is computed globally across all tokens.
This formulation enables ManualVLA to adaptively allocate computation according to the distinct characteristics of manual and action generation, while maintaining a unified architecture for the collaborative execution of final goal conditioned manipulation tasks.

\textbf{Action and Robot State Components.}
ManualVLA employs a diffusion-based approach for action modeling, therefore, a noise encoder is introduced to inject the noised actions into the action expert, and a noise decoder predicts the noise from the latent representations. Both modules are implemented as two-layer MLPs. In addition, the robot state is incorporated into the action expert through another two-layer MLP (state encoder), enabling the model to condition action generation on the current proprioceptive state.

\subsection{Manual and Action Generation via CoT}
\label{sec:4.3}

Given the final goal state of a task, ManualVLA generates subgoal manuals at key steps. 
To better produce corresponding action sequences based on the current observation and the generated manuals, we introduce a Manual Chain-of-Thought (ManualCoT) reasoning process including both explicit and implicit CoT.
Specifically, we introduce the details of manual generation in Section \ref{sec:4.3.1}, followed by the method for generating executable actions through ManualCoT in Sections \ref{sec:4.3.2}. Finally, in Sections \ref{sec:4.3.3}, we describe how both processes are unified within a single token sequence for the end-to-end training.

\subsubsection{Subgoal Manual Generation}
\label{sec:4.3.1}

To accomplish long-horizon manipulation tasks with predefined goal states, we design a multimodal manual that consists of (1) textual descriptions for subtask reasoning, (2) next-step images providing semantically rich conditioning, and (3) low-level position prompts for precise manipulation guidance.
As shown in Figure~\ref{fig:method} \textcolor{red}{(a)}, upon receiving the language instruction along with the current and final state images, ManualVLA first generates a textual component that describes object attributes and actions.
For target positions, we represent them using the pixel-level $(U, V)$ coordinates of each object's centroid.
Finally, subgoal image generation helps the model better model the physical world dynamics.

We assume that accomplishing a long-horizon task does not require extreme dense temporal subgoals.
Instead, providing guidance only at key frames where the task state changes is sufficient, such as when placing a pair of bricks onto the board.
ManualVLA generates a new manual only after the completion of the previous subgoal, ensuring efficient planning without redundant intermediate guidance. To achieve this, we first generate the text description in the manual. 
If the generated description of the manipulated objects differs from the previous planning output, ManualVLA proceeds to produce an entirely new manual. For example, the text output changes from “yellow blocks” to “purple blocks”. Otherwise, it reuses the previously generated manual for subsequent action generation.

\subsubsection{Manual-Conditioned Action Generation}
\label{sec:4.3.2}
Based on the generated subgoal manual, ManualVLA executes actions in a closed-loop manner, progressively generating the action sequence until the subgoal is achieved.
As shown in Figure~\ref{fig:method} \textcolor{red}{(a)}, leveraging the effectiveness of visual prompts for manipulation, we use the predicted $(U, V)$ coordinates to overlay a mask on the current image, highlighting the affordance region that serves as input to the action expert.
This construction of a prompt image to guide the action learning is defined as \textbf{explicit CoT reasoning}.
Meanwhile, we introduce an implicit CoT reasoning process within the shared attention module.
As shown in Figure~\ref{fig:method} \textcolor{red}{(b)}, in the latent space, the subgoal manual serves as a conditioning signal for action modeling through our constructed cross-task attention mask.
This conditioning information first informs the model about ``what'' object to manipulate, then specifies ``where'' the object should be placed, and finally provides the anticipated visual outcome after the manipulation. This CoT reasoning process performed in the latent space is referred to as \textbf{implicit CoT reasoning}.
By incorporating both explicit and implicit CoT reasoning processes, ManualVLA significantly improves its success rate in long-horizon tasks, as demonstrated in Section~\ref{sec: AB}.

\vspace{0.2cm}
\subsubsection{Token Sequence Design}
\label{sec:4.3.3}

After introducing the processes of manual and action generation, this section describes how ManualVLA jointly learns both tasks within a unified token sequence, while employing planning and action experts to specialize in different tasks. As shown in Figure~\ref{fig:method}, the language instruction, along with the current and goal scene images, is first inserted into the token sequence as the condition for subgoal manual generation. The subsequent tokens represent the generated manual, including object descriptions, target coordinates, and subgoal image, all of which are processed by the planning expert. Following this, the sequence incorporates the prompt image used in explicit CoT reasoning, as well as the robot state and noised action embeddings, which are handled by the action expert. As shown in Figure~\ref{fig:method} \textcolor{red}{(b)}, a cross-task shared attention mechanism is designed to allow the action expert to attend to the subgoal manual representations while masking out earlier inputs, thereby enabling effective information exchange between the two experts and fostering coherent reasoning across planning and action generation.

\vspace{-0.2cm}
\subsection{Training Strategy}
\label{sec:4.4}
As shown in Figure~\ref{fig:method} \textcolor{red}{(c)}, we train ManualVLA in three stages.
Before training, we initialize ManualVLA with the pretrained parameters of Janus-Pro \cite{chen2025janus}, and duplicate its LLM to separately initialize the planning and action experts.

\noindent\textbf{\textit{Stage 1: Action expert pretraining.}}
During this pretraining stage, we curated an assembly dataset by carefully filtering large-scale cross-embodiment datasets~\cite{open_x_embodiment_rt_x_2023, khazatsky2024droid, wu2024robomind} to update all parameters of the action expert.
As detailed in the Appendix~\ref{sec:appendix_data}, the resulting dataset comprises over 400K trajectory samples.
ManualVLA was trained on this dataset for five epochs, where the only conditioning inputs are the language instruction, a current scene image, and robot state. 
Following diffusion policy~\cite{chi2023diffusion}, the training objective is the mean squared error (MSE) between the predicted noises $\hat{\epsilon}^i$ at the $i$-th denoising steps and the ground-truth noises $\epsilon$, defined as:
$
\mathcal{L}_{\text{action}} = 
\mathbb{E}_{\epsilon \sim \mathcal{N}(0,1), i}
\left\| \hat{\epsilon}^i - \epsilon \right\|_2^2.
$
\noindent\textbf{\textit{Stage 2: Manual expert pretraining.}}
At this stage, we train only the manual expert using data synthesized with our digital-twin toolkit, resulting in a dataset of over 10K frames for each task.
Following Janus-Pro~\cite{chen2025janus}, the model is supervised with a cross-entropy loss $\mathcal{L}_{\text{manual}}$ applied to the subgoal manual, which includes the object description, target position, and subgoal image tokens.

\noindent\textbf{\textit{Stage 3: Joint manual-action fine-tuning.}}
\label{sec:training_strategy}
Benefiting from the stable generation capabilities acquired during pretraining, we collect 100 demonstrations for each downstream task using master-puppet teleoperation. Objects are placed at diverse locations on the table to ensure sufficient variation. Each demonstration includes both action execution data and automatically extracted manual data.
All components of ManualVLA are then jointly trained using the token sequences defined in Section~\ref{sec:4.3.1}.
The final objective is defined as:
$
\mathcal{L}_{\text{final}} = \mathcal{L}_{\text{manual}} + \mathcal{L}_{\text{action}}.
$

\begin{figure}[h]
    \centering
    \includegraphics[width=1.0\linewidth]{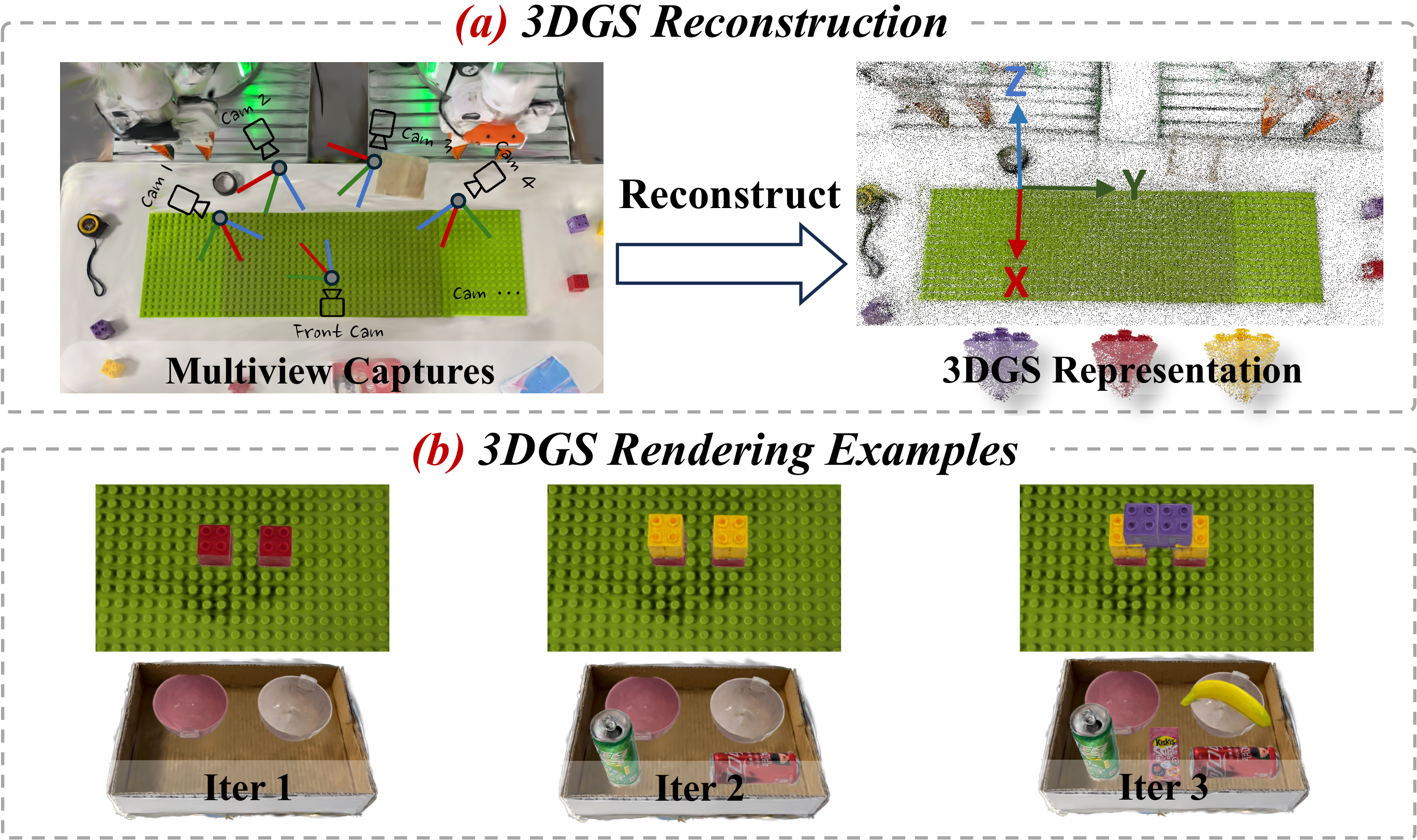}
    \caption{\textbf{Digital-twin example.} 
  \textcolor{red}{(a)} We reconstruct 3D Gaussian Splatting representations, which are then decomposed into the LEGO board and individual bricks.
\textcolor{red}{(b)} We iteratively place the bricks on the board or objects on the box.}
    \label{fig: sim_to_real}
    \vspace{-0.55cm}
\end{figure}

\subsection{Real-to-Render Data Generation}
\label{sec:3.5}

To construct downstream task data, we develop a high-fidelity digital-twin toolkit based on 3D Gaussian Splatting~\cite{2023_8_08-3dgs_for_real_time_radiance_field_rendering} to automatically generate training data with intermediate target states. For example, in Figure~\ref{fig: sim_to_real}~\textcolor{red}{(a)}, we first reconstruct 3D assets of the LEGO board and individual bricks through multi-view capture, and then align them to a unified Cartesian coordinate system for consistent spatial referencing. Subsequently, we follow an iterative placement procedure as shown in Figure~\ref{fig: sim_to_real}~\textcolor{red}{(b)}, where given an initial state and a set of available bricks, we sequentially place each brick by randomly sampling a valid position on the board. At each intermediate state, we render the current configuration from a front-view camera perspective. This process produces photorealistic images for each assembly or rearrangement step, along with position and textual information.
More details are provided in Appendix~\ref{sec:appendix_data}.

\section{Experiments}

We organize our experiments as follows. Section \ref{sec: ES} defines the tasks and introduces the baseline methods. Section \ref{sec: MR} compares ManualVLA with these baselines on manual generation and manipulation performance. In Section \ref{sec: AB}, we assess the effectiveness of our MoT architecture and CoT reasoning mechanism. Finally, Section \ref{sec:GA} evaluates the generalization of ManualVLA to unseen object shapes, backgrounds, and lighting conditions. All the above experiments are performed on a dual-arm Franka robotic platform. Additionally, we validate the advantages of our method on general manipulation tasks on RLBench~\cite{james2020rlbench} Benchmark in Section~\ref{sec: SE}.

\subsection{Experimental Setup}
\label{sec: ES}
\subsubsection{Task Definition}
We design three long-horizon tasks with defined goal states, challenging the model's procedural reasoning and manipulation capabilities.
\textbf{(1) 2D LEGO Assembly:} The task begins with several LEGO bricks of different colors placed on a planar board. 
Given the final 2D assembled structure as the goal, the model must infer a sequence of intermediate manipulation actions and execute them through coordinated bimanual control.
\textbf{(2) 3D LEGO Assembly:} The task extends the 2D LEGO Assembly task to a more challenging 3D setting, where the final configuration transitions from a planar layout to a 3D structure. This upgraded task imposes greater demands on the model's spatial reasoning abilities.
\textbf{(3) Object Rearrangement:} The task begins with several objects of diverse shapes, sizes, and semantics scattered around a box. 
Given a goal state in which all objects are placed at their designated positions inside the box, the model must progressively generate manipulation actions, alternating control of the left and right arms to prevent collisions.

\subsubsection{Baselines}
We compare ManualVLA against three categories of strong VLA baselines that represent the state of the art in robotic manipulation.
\textbf{(1) First category:}
$\pi_0$~\cite{black2024pi_0}, $\pi_{0.5}$~\cite{intelligence2025pi05visionlanguageactionmodelopenworld}, and FAST~\cite{pertsch2025fast}, which adopt robust action-generation paradigms.
We load the official pretrained weights provided by each method and strictly follow their fine-tuning protocols, except that we additionally embed the final goal image into the model as the target condition.
\textbf{(2) Second category:}
CoT-VLA~\cite{zhao2025cot}, which not only incorporates the final goal image as an additional condition, but also predicts key subgoal future images.
For fair comparison, we align the supervision of subgoal image generation with that used in ManualVLA.
\textbf{(3) Third category:}
We introduce a hierarchical baseline that combines a VLM~\cite{chen2025janus} with \textbf{$\pi_{0.5}$}.
The VLM is trained to generate visual and language prompts similar to our method, while \textbf{$\pi_{0.5}$} is trained to interpret these prompts and generate actions.
This baseline can be regarded as a hierarchical variant of our manual-generation approach.

\subsection{Main Results}
\label{sec: MR}

\subsubsection{Manual Generation}
\label{sec: MG}

\begin{table}[h]
  \centering
  \caption{Quantitative results of ManualVLA in generating subgoal images and ($U,V$) coordinates across the three downstream tasks.}
  \label{tab:manual_results}
  \vspace{-5pt}
  {\scriptsize
  \renewcommand{\arraystretch}{0.9} 
  \resizebox{\columnwidth}{!}{
  \begin{tabular}{lccc}
    \toprule
    \multirow{2}{*}{\textbf{Task}} & \multicolumn{2}{c}{\textbf{Subgoal Image}} & \textbf{\textit{(U,V)}} \\
    \cmidrule(lr){2-3} \cmidrule(lr){4-4}
     & PSNR $\uparrow$ & FID $\downarrow$ & MAE $\downarrow$ \\
    \midrule    
    2D LEGO Assembly & 29.01 & 36.39 & 3.23\\
    3D LEGO Assembly & 28.68 & 34.63 & 3.58\\
    Object Rearrangement & 28.11 & 24.46 & 6.21\\
    \bottomrule
  \end{tabular}
  }
  }
   \vspace{-0.25cm}
\end{table}

\begin{table*}[t]
\centering
\setlength{\tabcolsep}{1.5pt}
{\scriptsize 
\caption{
\textbf{Comparison of ManualVLA and baselines.}
We report the manipulation success rate (S.R.) for the complete long-horizon tasks using 20 unseen test goal states, and additionally report the success rate of key intermediate steps.
}
\vspace{-0.3cm}
\label{tab:main}

\resizebox{\textwidth}{!}{%
\begin{tabular}{l|cccc|cccc|cccc}
\toprule
\multirow{2}{*}{Method} &
\multicolumn{4}{c|}{\textbf{2D LEGO Assembly}} &
\multicolumn{4}{c|}{\textbf{3D LEGO Assembly}} &
\multicolumn{4}{c}{\textbf{Object Rearrangement}}\\

\cmidrule(lr){2-5}
\cmidrule(lr){6-9}
\cmidrule(lr){10-13}

 & 2 bricks $\rightarrow$ & 2 bricks $\rightarrow$ & 2 bricks $\rightarrow$ & \textbf{S.R.}
 & 2 bricks $\rightarrow$ & 2 bricks $\rightarrow$ & 2 bricks $\rightarrow$ & \textbf{S.R.}
 & 2 objects $\rightarrow$ & 2 objects $\rightarrow$ & 2 objects $\rightarrow$ & \textbf{S.R.} \\
\midrule

$\pi_{0}$~\cite{black2024pi_0} &0.25&0.20&0.15&0.15&0.25&0.15&0.10&0.10&0.35&0.20&0.10&0.10\\

$\pi_{0.5}$~\cite{intelligence2025pi05visionlanguageactionmodelopenworld} &
0.30&0.25&0.25&0.20&0.30&0.20&0.15&0.15&0.45&0.25&0.15&0.15\\

FAST~\cite{pertsch2025fast} &
0.20&0.15&0.10&0.10&0.15&0.15&0.05&0.05&0.20&0.15&0.05&0.05\\

CoT-VLA~\cite{zhao2025cot} &
0.40&0.35&0.30&0.30&0.35&0.30&0.25&0.25&0.60&0.40&0.30&0.30\\

VLM + $\pi_{0.5}$ &
0.75 & 0.70 & 0.65 & 0.60 &
0.65 & 0.45 & 0.35 & 0.35 &
\textbf{0.90} & 0.55 & 0.65 & 0.50 \\

\textbf{ManualVLA} &
\textbf{0.95} & \textbf{0.90} & \textbf{0.85} & \textbf{0.85} &
\textbf{0.90} & \textbf{0.75} & \textbf{0.65} & \textbf{0.65} &
\textbf{0.90} & \textbf{0.70} & \textbf{0.80} & \textbf{0.65} \\

\bottomrule
\end{tabular}
}
}
 \vspace{-0.35cm}
\end{table*}

We first evaluate the capability of the planning expert in ManualVLA to generate high-fidelity manuals on 300 unseen test samples. As shown in Table~\ref{tab:manual_results}, our model produces satisfactory intermediate images across all three tasks, achieving high PSNR scores, indicating strong structural and pixel-level consistency with the ground truth. Furthermore, the low FID scores, particularly in the Object Rearrangement task, demonstrate that the generated image distribution closely matches that of real images, confirming their realism and fidelity. The remarkably low MAE scores further highlight ManualVLA's precision in predicting the position of target objects. For language descriptions, we evaluate accuracy by checking the predicted object nouns, all of which are correctly generated on unseen test samples.

\begin{figure}[t]
\includegraphics[width=0.47\textwidth]{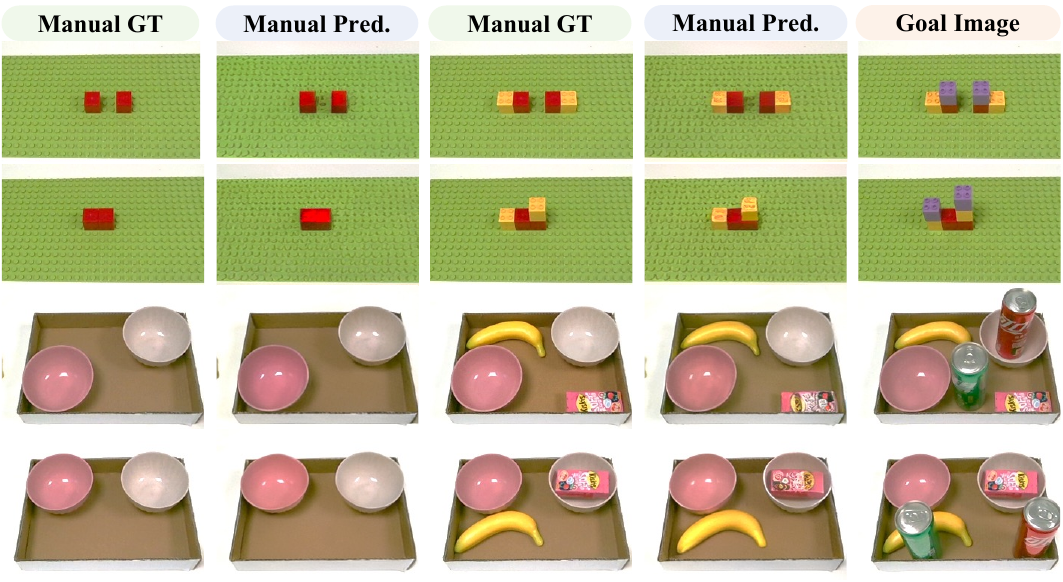}
\centering
\caption{For each task, we visualize three components: (1) manual ground truth (GT), (2) manual predictions (Pred.) generated by ManualVLA, and (3) the final goal image.
}
\label{fig:vis}
\vspace{-0.1cm}
\end{figure}

\begin{figure}[t]
\includegraphics[width=0.47\textwidth]{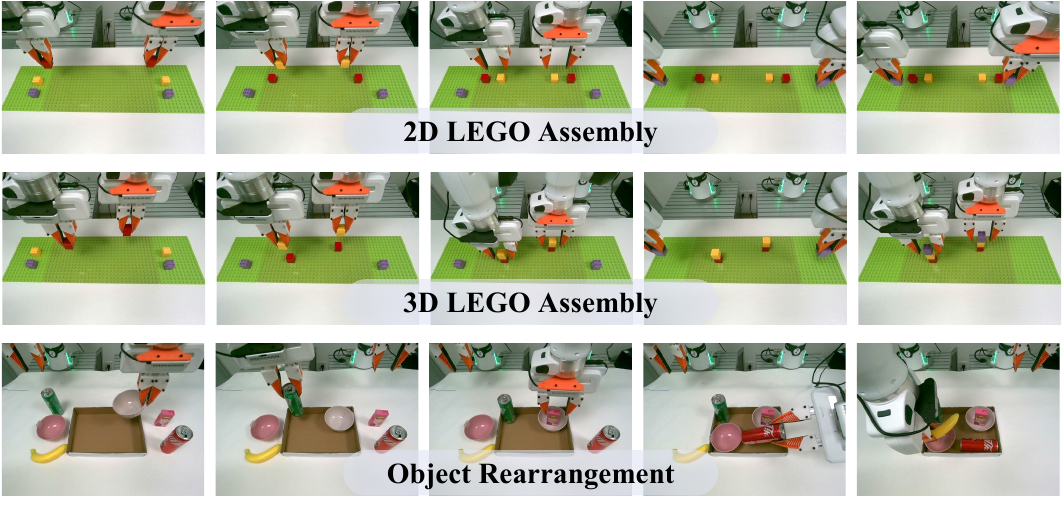}
\centering
\caption{Visualization of real-world experiments on Franka Research 3 dual-arm robots, executed from left to right.}
\label{fig:actionvis}
\vspace{-0.4cm}
\end{figure}

Figure \ref{fig:vis} presents qualitative results of ManualVLA on 2D and 3D LEGO assembly as well as object rearrangement tasks. The model accurately predicts intermediate steps that align with the ground truth, effectively capturing both spatial arrangements and object identities. In LEGO assembly, ManualVLA sequentially reconstructs the correct brick placements and colors, demonstrating precise step-wise reasoning. 
For object rearrangement, it gradually progresses toward the final goal configuration and accurately generates the spatial relationships between objects.
Overall, these results highlight ManualVLA's strong intermediate reasoning capabilities in long-horizon tasks, establishing a reliable foundation for the action expert to generate accurate actions.

\vspace{-0.16cm}
\subsubsection{Action Generation}
\label{sec: AG}
Across all three real-world long-horizon tasks, ManualVLA achieves the highest success rates, markedly outperforming all baselines. As shown in Table \ref{tab:main}, we report both step-wise subgoal accuracy and end-to-end task success. Compared with the strongest hierarchical baseline, ManualVLA improves the final task completion rate by 15\%-30\%. While baseline models often succeed in the early stages of a long-horizon pipeline, they typically fail to sustain this performance through the whole sequence. In contrast, ManualVLA mitigates this degradation by decomposing complex tasks into structured subgoal manuals and grounding them into precise actions through a combination of explicit and implicit reasoning, enabling consistent performance throughout the entire task.
Note that, as reflected in the manual generation results, the generated manuals may contain minor inaccuracies. Nevertheless, with the ManualCoT strategy and the capacity of our MoT architecture, ManualVLA remains robust and can still produce reliable actions even under moderate manual errors.
In Appendix~\ref{sec:appendix_experiments}, we also validate the advantages of our method on general manipulation tasks.

The qualitative rollouts in Figure \ref{fig:actionvis} further corroborate these results. ManualVLA generates structured and interpretable intermediate states that reliably guide the dual-arm system through precise grasping and relocation motions. In both LEGO assembly tasks, the robot maintains accurate brick alignment across all stages, while in the object-rearrangement task, it robustly manipulates objects with varying shapes, textures, and occlusions. More visualizations, failure case analyses, and execution videos are provided in Appendix~\ref{sec:appendix_visualizations}, Appendix~\ref{sec:appendix_failure}, and the supplementary material.

\begin{figure*}[t]
    \centering
    \includegraphics[width=1.0\textwidth]{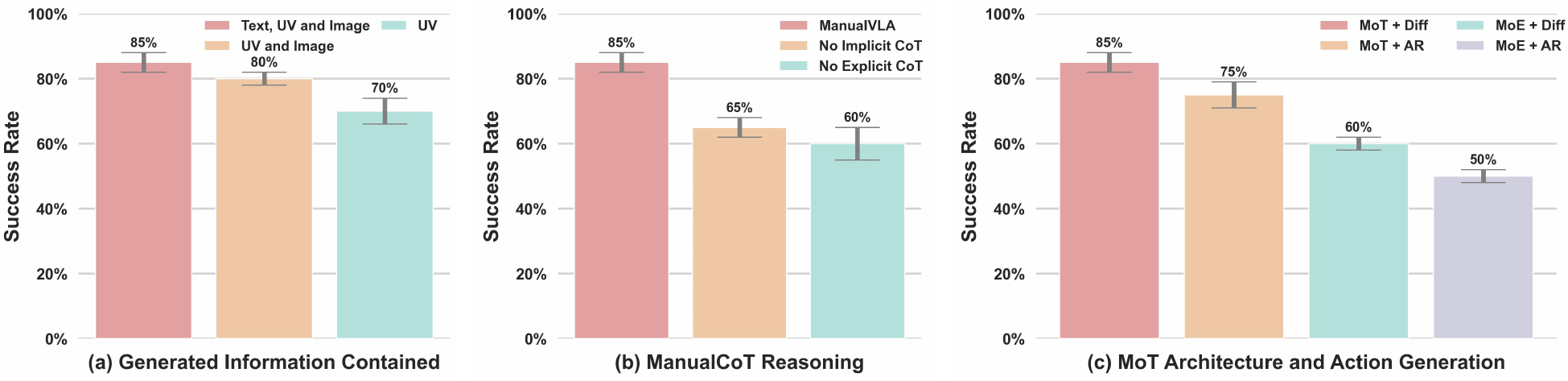} 
    \vspace{-0.6cm}
    \caption{\textbf{Ablation study.} We investigate the impact of (a) the information contained in the generated manuals, (b) explicit and implicit CoT reasoning, (c) the MoT architecture design, and (d) the action generation paradigm on long-horizon manipulation success rates.}
    \label{fig:abla}
    \vspace{-0.2cm}
\end{figure*}

\subsection{Ablation Study}
\label{sec: AB}
We conduct detailed ablation studies on the 2D LEGO Assembly task, reporting the long-horizon task success rate.
\textbf{(a) What information should a generated manual contain?}
We explore three variants: generating only the target positions ($U,V$), generating both target positions and subgoal images, and generating the full set of multimodal manual.
As shown in Figure~\ref{fig:abla} (a), increasing the amount of multimodal information in the manuals leads to improved manipulation performance.
Note that in this experiment, we consistently use explicit CoT visual-prompt images as inputs to the action expert.
The results demonstrate that high-level textual descriptions for subtask understanding, next-step images for semantically reasoning, and position prompts for accurate localization all serve as critical implicit conditions that enable precise manipulation.
\textbf{(b) Importance of Explicit CoT and Implicit CoT.}
We examine two variants of our ManualCoT reasoning:
(1) No Explicit CoT, where the action expert receives only the latent features from the planning expert together with the current image; and
(2) No Implicit CoT, where the action expert receives only the visual-prompt image.
As shown in Figure~\ref{fig:abla} (b), both variants lead to a noticeable performance degradation compared to ManualVLA, demonstrating that explicit and implicit CoT reasoning are jointly indispensable for solving long-horizon, goal-defined manipulation tasks.
\textbf{(c) The MOT Architecture and Action Generation Design.}
In Figure~\ref{fig:abla} (c), we compare our MoT architecture with a standard Mixture-of-Experts (MoE) architecture (duplicate only FFNs in LLM)~\cite{deng2025emerging}.
The results show that using an MoE strategy fails to produce high-quality manuals and actions simultaneously, both of which are crucial for long-horizon tasks.
Meanwhile, we find that for precise manipulation tasks, diffusion-based action generation yields superior performance. 
More ablations on how manual quality influences manipulation performance are provided in Appendix~\ref{sec:appendix_experiments}.

\subsection{Generalization Analysis}
\label{sec:GA}
First, in Section~\ref{sec: MR}, all final goal states used for evaluation differ from those in the teleoperated manipulation training set. Therefore, our main results simultaneously validate the generalization capability of ManualVLA with respect to both final goal states and object positions.
In addition, we evaluate the robustness of ManualVLA under variations in background, object shape, and lighting. As shown in Table~\ref{tab:gen}, these unseen perturbations are introduced in the 2D LEGO Assembly task and differ from all configurations seen during training. ManualVLA exhibits only a modest performance drop, which can be attributed to the rich guidance provided by our proposed manual generation expert and the CoT reasoning strategy during action prediction.
Moreover, our proposed digital-twin toolkit provides large-scale manual generation data, allowing the model to produce accurate manuals even in unseen scenarios.

\begin{table}[t]
\centering
\setlength\tabcolsep{8.5pt}
\caption{\textbf{Generalization.} We report the mean success rate and performance degradation ratio for each task over 20 rollouts under variations in background, object shape, and lighting.
}
\vspace{-0.3cm}
\includegraphics[width=0.48\textwidth]{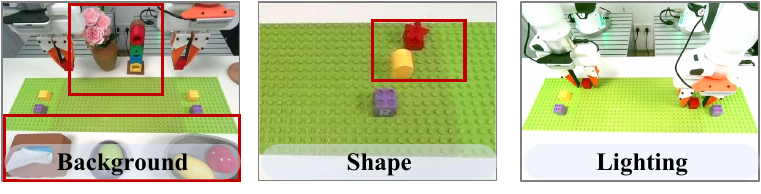}
\scriptsize
\begin{tabular}{c|cccc}
\toprule
2D LEGO& Origin & Background & Shape & Lighting  \\
\midrule
VLM + $ \pi_{0.5}$& 0.60& 0.45(-25\%) & 0.35(-46\%)& 0.50(-17\%)\\
ManualVLA& 0.85 & 0.65(-23\%)  & 0.60(-29\%) & 0.70(-17\%) \\
\bottomrule
\end{tabular}
\label{tab:gen}
\end{table}

\subsection{Simulation Experiment}
\label{sec: SE}

We evaluate ManualVLA on the RLBench~\cite{james2020rlbench} benchmark, comparing it against state-of-the-art (SOTA) VLA baselines. The results demonstrate ManualVLA's robust capabilities in predicting future images and generating precise actions.

\subsubsection{Simulation benchmark. }

\begin{table*}[t]
\caption{
\textbf{Comparison of ManualVLA and baselines on RLBench.}
We train all methods in the multi-task setting~\citep{shridhar2022peract} and report the success rates (S.R.) and variances (Var.).
}
\centering
\small
\resizebox{\textwidth}{!}{
\begin{tabular}{>{\raggedright\arraybackslash}p{3.3cm}|cccccccccc|c}
\toprule
\multirow{2}{*}{Models} & Close & Close & Toilet & Sweep & Close & Phone & Umbrella & Frame & Wine at & Water & Mean \\
 & box & laptop lid & seat down & to dustpan & fridge & on base & out & off hanger & rack & plants & S.R. \& Var  \\
\midrule
FAST~\cite{pertsch2025fast} & 0.85 & 0.70 & \textbf{0.90} & 0.45 & 0.60 & 0.15 & 0.15 & 0.25 & 0.45 & 0.15 & 0.47 \textcolor{gray}{$\pm0.03$} \\
$\pi_{0}$~\cite{black2024pi_0} & 0.85 & \textbf{0.95} & \textbf{0.90} & 0.85 & 0.80 & 0.25 & 0.20 & 0.65 & 0.65 & 0.25 & 0.63 \textcolor{gray}{$\pm0.01$} \\
$\pi_{0.5}$~\cite{intelligence2025pi05visionlanguageactionmodelopenworld} & \textbf{0.90} & 0.80 & \textbf{0.90} & 0.50 & 0.60 & 0.15 & 0.25 & 0.35 & \textbf{0.75} & \textbf{0.35} & 0.56 \textcolor{gray}{$\pm0.03$} \\   
CoT-VLA~\cite{zhao2025cot} & \textbf{0.90} & 0.85 & \textbf{0.90} & 0.60 & 0.70 & 0.20 & 0.30 & 0.55 & 0.55 & 0.30 & 0.59 \textcolor{gray}{$\pm0.03$}  \\
\rowcolor[HTML]{FFF0F5}
\textbf{ManualVLA~(ours)} & \textbf{0.90} & 0.90 & \textbf{0.90} & \textbf{1.00} & \textbf{0.85} & \textbf{0.30} & \textbf{0.50} & \textbf{0.70} & 0.65 & \textbf{0.35} & \textbf{0.70} \textcolor{gray}{$\pm0.02$}  \\

\bottomrule
\end{tabular}}
\label{tab:rlbench}
\end{table*}

To assess the fundamental manipulation capabilities of our method across common manipulation tasks, we conduct experiments on 10 tasks in the RLBench~\cite{james2020rlbench} benchmark based on the CoppeliaSim simulator. The task suite includes \textit{Close box}, \textit{Close Laptop}, \textit{Toilet seat down}, \textit{Sweep to dustpan}, \textit{Close fridge}, \textit{Phone on base}, \textit{Take umbrella out}, \textit{Take frame off hanger}, \textit{Place wine at rack}, and \textit{Water plants}. All tasks are executed on a Franka Panda robot equipped with a front-view RGB camera to get the visual input. We collect the data by following pre-defined waypoints and utilizing the Open Motion Planning Library~\cite{sucan2012open}. Building upon the frame-sampling technique employed in previous studies~\cite{shridhar2022peract, goyal2023rvt, jia2025lift3d}, we construct a training dataset where each task contains 100 trajectories.
To generate ground-truth $(U,V)$ labels, we follow the key-frame extraction procedure from prior work. Specifically, we extract the end-effector poses of the key frames and then use the camera parameters to project their world-coordinate positions into $(U,V)$ coordinates.

\subsubsection{Training and evaluation details. }

We compare ManualVLA against four state-of-the-art (SOTA) VLA models, including FAST~\cite{pertsch2025fast}, $\pi_0$~\cite{black2024pi_0}, $\pi_{0.5}$~\cite{intelligence2025pi05visionlanguageactionmodelopenworld}, and CoT-VLA~\cite{zhao2025cot}. While the former three adopt robust action-generation paradigms, CoT-VLA conditions on the final goal image and additionally predicts future subgoal images. Specifically, FAST~\cite{pertsch2025fast} utilizes autoregressive action outputs, $\pi_0$~\cite{black2024pi_0} and $\pi_{0.5}$~\cite{intelligence2025pi05visionlanguageactionmodelopenworld} employ flow matching, and CoT-VLA~\cite{zhao2025cot} combines autoregressive image generation with diffusion-based action prediction. For all baselines, we initialize with the official pretrained parameters and strictly adhere to their original fine-tuning configurations. To ensure a fair comparison, we align the subgoal supervision in CoT-VLA to match the formulation used in ManualVLA.
For ManualVLA’s input, the single-view RGB image is resized to $384 \times 384$, with text instructions derived directly from the simulation environment and the robot state is aligned with the predicted actions. 
ManualVLA model is trained for 500 epochs using the AdamW optimizer~\cite{loshchilov2017decoupled} and CosineAnnealingLR~\cite{loshchilov2017sgdrstochasticgradientdescent} on 8 NVIDIA H20 GPUs, with mixed-precision training employed.
Following~\cite{li2024cogact, goyal2023rvt}, we evaluate all methods using 20 rollouts from the latest epoch checkpoint, repeating the evaluation three times for each task and reporting the average success rate along with the variance.

\subsubsection{Quantitative results. }

As presented in Table~\ref{tab:rlbench}, ManualVLA achieves an average success rate of 70\% across 10 diverse tasks, surpassing the previous SOTA methods $\pi_0$~\cite{black2023zero} and CoT-VLA~\cite{zhao2025cot} by margins of 7\% and 11\%, respectively. 
Specifically, ManualVLA attains superior performance on 8 out of 10 tasks, highlighting the advantage of ManualCoT strategy in guiding precise action generation. By generating sub-goal images and constructing visual prompt images, ManualVLA effectively leverages the fine-grained affordance guidance provided by explicit CoT reasoning. Furthermore, the MoT architecture, equipped with a shared attention module, enables robust task understanding and action generation conditioned on the sub-goal manual within the latent space. Through the integration of both explicit and implicit CoT reasoning, ManualVLA demonstrates substantial improvements in tasks requiring precise actions, such as \textit{sweep to dustpan} and \textit{take out umbrella}, compared to $\pi_0$ and $\pi_{0.5}$.

\vspace{-0.1cm}
\section{Conclusion}
In this work, we address the challenge of enabling robots to autonomously perform long-horizon tasks with defined goal states, such as LEGO assembly and object rearrangement. To this end, we introduce ManualVLA, a unified VLA model built on a Mixture-of-Transformers architecture that couples multimodal manual generation with action execution. Central to the model is a Manual Chain-of-Thought (ManualCoT) process, which converts subgoal manuals into precise actions by using them as explicit control conditions and implicit manipulation cues. To support scalable training, we further develop a 3D Gaussian Splatting-based digital-twin pipeline that automatically produces large amounts of manual data.
Experimental results on long-horizon tasks show that ManualVLA achieves a 32\% higher average success rate than existing VLA methods.
{
    \small
    \bibliographystyle{ieeenat_fullname}
    \bibliography{main}
}

\clearpage
\setcounter{page}{1}

\appendix

\section{Additional Dataset Details}
\label{sec:appendix_data}

\begin{figure}[t]
    \centering
    \vspace{-0.2cm}
    \includegraphics[width=0.47\textwidth]{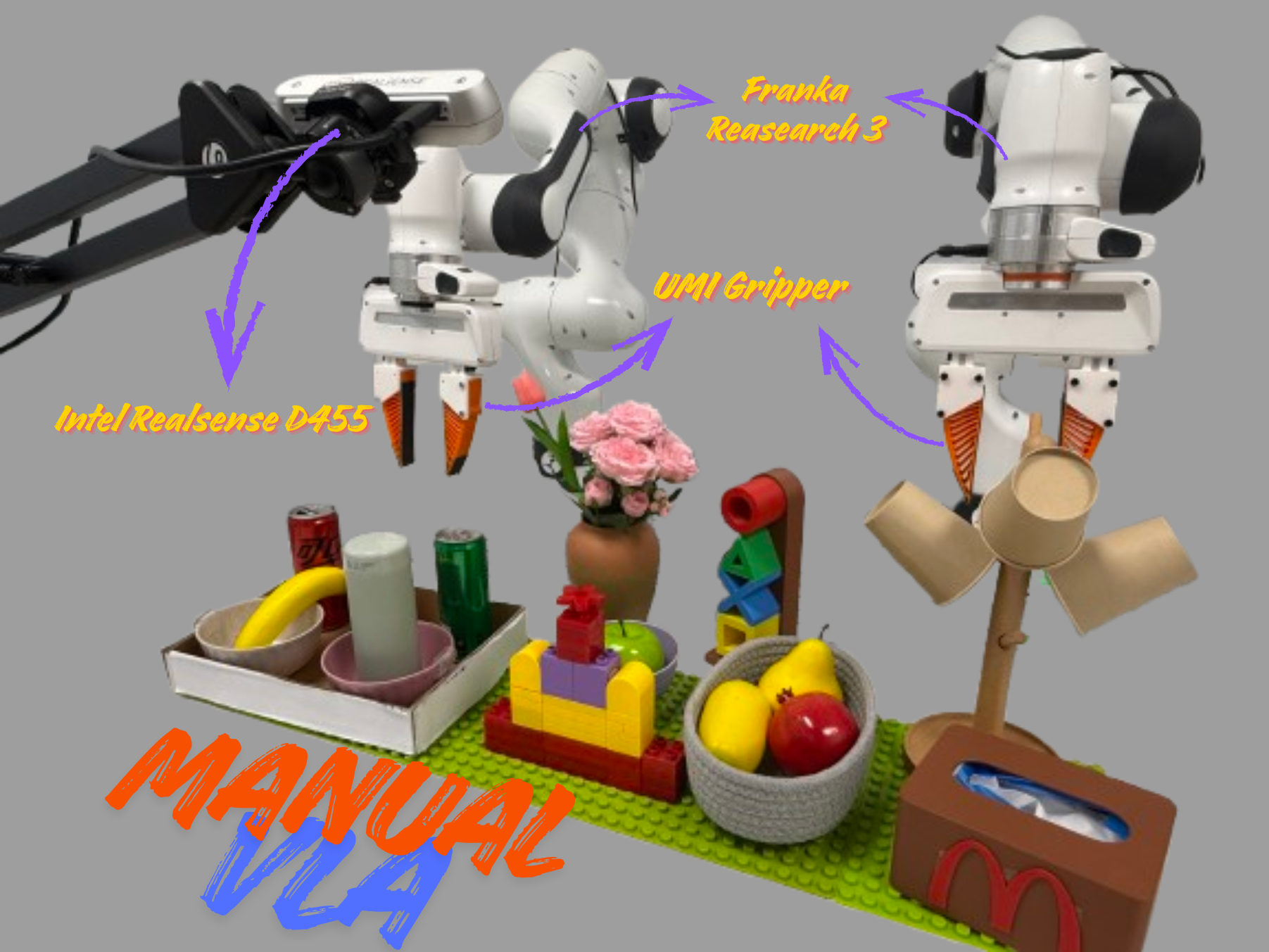} 
    \vspace{-0.3cm}
    \caption{\textbf{Real-World Assets and Experimental Settings.} We provide visualizations of the assets used and the hardware settings for Dual-Arm Franka Platform tasks. }
    \label{fig:asset}
    \vspace{-0.3cm}
\end{figure}

This section provides further details on our pretraining and real-world datasets used in the multi-stage training process. Furthermore, we elaborated in detail on the digital-twin data generation pipeline.

\subsection{Pretraining Data Curation}
\label{sec:appendix_pretraining}

As outlined in Stage 1 of our training strategy, the action expert is pretrained on a large-scale, cross-embodiment assembly dataset. To build this dataset, we carefully aggregate and curate demonstrations from public robotic manipulation datasets, including Open X-Embodiment~\citep{open_x_embodiment_rt_x_2023}, Droid~\citep{khazatsky2024droid}, and Robomind~\citep{wu2024robomind}. 
These datasets contain millions of trajectories from a diverse set of robotic platforms, sensor configurations, and task settings. However, only a subset of these demonstrations is directly relevant to the domain of robotic assembly and rearrangement. Therefore, a careful multi-stage filtering and unification process is required before pretraining. We first apply a task-level filtering pipeline designed to extract trajectories involving low-level manipulation primitives that commonly appear in assembly and rearrangement settings, such as precise object-manipulation, and general pick–place.
After filtering and integration, the assembly pretraining dataset contains more than 400,000 high-quality trajectory samples, each representing a structured demonstration of object manipulation. Although this is a fraction of the raw data available in the source corpora, the curated subset is specifically optimized for assembly-centric skills while remaining a robust foundation for the action expert to learn a wide range of manipulation primitives before fine-tuning on downstream tasks.

\begin{table}[h]
    \caption{
    \textbf{The dataset name and sampling weight used in our mixed large-scale pretraining dataset.}
    }
    \centering
       \begin{tabular}{lr}
        \toprule
        \multicolumn{2}{c}{\textbf{Training Dataset Mixture}}\\
        \midrule
        Fractal~\citep{brohan2022rt} & 6.8\% \\
        Kuka~\citep{kalashnikov2018qt} & 10.5\% \\
        Bridge\citep{ebert2021bridge, walke2023bridgedata} & 4.9\% \\
        Taco Play~\citep{rosetebeas2022latent,mees2023grounding} & 2.5\% \\
        Jaco Play~\citep{dass2023jacoplay} & 0.4\% \\
        Berkeley Cable Routing~\citep{luo2023multistage} & 0.2\% \\
        Roboturk~\citep{DBLP:journals/corr/abs-1811-02790} & 2.0\% \\
        Viola~\citep{zhu2023viola} & 0.8\% \\
        Berkeley Autolab UR5~\citep{BerkeleyUR5Website} & 1.0\% \\
        Toto~\citep{zhou2023train} & 1.7\% \\
        Language Table~\citep{lynch2023interactive} & 3.7\% \\
        Stanford Hydra Dataset~\citep{belkhale2023hydra}  & 3.8\% \\
        Austin Buds Dataset~\citep{zhu2022bottom}  & 1.8\% \\
        NYU Franka Play Dataset~\citep{cui2022play}  & 0.7\% \\
        Furniture Bench Dataset~\citep{heo2023furniturebench}  & 2.1\% \\
        UCSD Kitchen Dataset~\citep{ucsd_kitchens}  &  $<$0.1\% \\
        Austin Sailor Dataset~\citep{nasiriany2022sailor}  & 1.9\% \\
        Austin Sirius Dataset~\citep{liu2022robot}  & 1.5\% \\
        DLR EDAN Shared Control~\citep{quere_shared_2020}  & $<$0.1\% \\
        IAMLab CMU Pickup Insert~\citep{saxena2023multiresolution}  & 0.7\% \\
        UTAustin Mutex~\citep{shah2023mutex} & 1.9\% \\
        Berkeley Fanuc Manipulation~\citep{fanuc_manipulation2023} & 0.6\% \\
        CMU Stretch~\citep{mendonca2023structured} & 0.1\% \\
        BC-Z~\citep{jang2022bc} & 6.3\% \\
        FMB Dataset~\citep{luo2024fmb}  & 6.0\% \\
        DobbE~\citep{shafiullah2023dobbe}  & 1.2\% \\
        DROID~\citep{khazatsky2024droid}  & 14.2\% \\
        Stanford Kuka Dataset~\cite{lee2019makingsensevisiontouch} & 0.3\%\\
        Stanford Robocook Dataset~\cite{shi2023robocooklonghorizonelastoplasticobject} & 0.2\% \\
        Columbia Cairlab Pusht Real~\cite{chi2023diffusion} & $<$0.1\% \\
        UCSD Pick and Place & 0.8\% \\
        Maniskill~\cite{gu2023maniskill2unifiedbenchmarkgeneralizable} & 7.5\% \\
        Berkeley RPT~\cite{radosavovic2023real} & $<$0.1\% \\
        QUT Dexterous Manipulation~\cite{ceola2023lhmanip} & $<$0.1\% \\
        RoboSet~\cite{kumar2023robohive} & 5.2\% \\
        BridgeData V2~\cite{walke2023bridgedata} & 9.3\% \\
        RoboMind~\cite{wu2024robomind} & 1.2\% \\
        \bottomrule
        \end{tabular}%
        \label{tab:data_mix}
\end{table}

\subsection{Real-World Data Collection}
\label{sec:appendix_real_world_data}

For real-world experiments, we evaluate three downstream tasks (2D LEGO Assembly, 3D LEGO Assembly, Object Rearrangement) on the dual-arm Franka platform. Below, we detail the hardware configurations, task settings, and data protocols.

\textbf{Hardware Configurations.} We equip the dual-arm Franka experimental environment with two Franka Research 3 arms each with a 3D-printed UMI gripper, the configurations of which is summarized in Table~\ref{tab:franka_setup}. As shown in Figure~\ref{fig:asset}, we utilize an Intel RealSense 455 camera to capture a static third-person view at the speed of 30Hz. 

\begin{table}[h]
\centering
\caption{\textbf{The hardware setups of the Franka Research 3, including joint position limits and velocity limit. }}
\label{tab:franka_setup}
\begin{tabular}{ccc}
\hline
Joint Number & Position Limits & Velocity Limits \\
\hline
J1 & $-166\degree \sim +166\degree$ & $150\degree/s$ \\
J2 & $-105\degree \sim +105\degree$ & $150\degree/s$ \\
J3 & $-166\degree \sim +166\degree$ & $150\degree/s$ \\
J4 & $-176\degree \sim -7\degree$ & $150\degree/s$ \\
J5 & $-165\degree \sim +165\degree$ & $301\degree/s$ \\
J6 & $+25\degree \sim +265\degree$ & $301\degree/s$ \\
J7 & $-175\degree \sim +175\degree$ & $301\degree/s$ \\
\hline
\end{tabular}
\end{table}

\textbf{Task Settings.} We provide a detailed explanation of the assembly and rearrangement tasks and their success conditions. \textbf{\textit{2D Assembly:}} The final state of the task is that LEGO blocks of different colors are randomly inserted into the same layer position on the planar board, without any stacking of blocks. Given the final 2D assembled structure as the goal, the robot arms need to pick up the LEGO blocks from both sides of the board in turn and insert them into the correct corresponding position in the center of the board. We only consider it a success if the current Lego blocks on the board match the position of the final structure at each key intermediate step. For a complete evaluation, only if all key intermediate steps are successful will it be counted as successful. \textbf{\textit{3D Assembly:}} Based on the 2D Assembly Task, the 3D Assembly Task allows for more complex placement situations such as stacking between LEGO blocks. For evaluation, we do not require the placement order of LEGO blocks, but we still require that the LEGO blocks placed at each key intermediate step conform to the placement position of the corresponding color LEGO block in the final state, and the final 3D LEGO shape needs to match the given shape. \textbf{\textit{Object Rearrangement:}} Unlike Assembly tasks, we use bowls, bananas, beverage cans and other common objects in life. The goal of the task is to pack the objects on the table into the box in turn and conform to the given placement pattern. In the rearrangement task, the order of placement is very important since there should be no situation where the objects in the bowl are placed before the bowl during execution. We follow the same evaluation settings as Assembly tasks at key intermediate steps and final states.

\textbf{Data Protocols.} For each task, we collect 100 demonstrations using 3DConnexion Spacemouse to teleoperate each Franka arm with target positions randomized on the table and box to promote data diversity. Language instructions are manually created and diversified via augmentation. Each trajectory is recorded at a frequency of 15hz, and each step contains a third-angle image shaped 640x480, the dual-arm end effector poses and the grippers discrete opening and closing. For each trajectory, we automatically filter key intermediate steps by grippers status to obtain the subgoal image, and use pixel matching to obtain the low-level $(U, V)$ coordinates corresponding to each image.

\subsection{Digital-Twin Data Generation}
\label{sec:appendix_digital_twin}

To train the planning expert (Stage 2) without the prohibitive cost of large-scale human annotation, we developed a high-fidelity digital-twin toolkit based on 3D Gaussian Splatting (3DGS). The pipeline consists of two main steps:

\begin{figure*}[t]
\centering
\includegraphics[width=\textwidth]{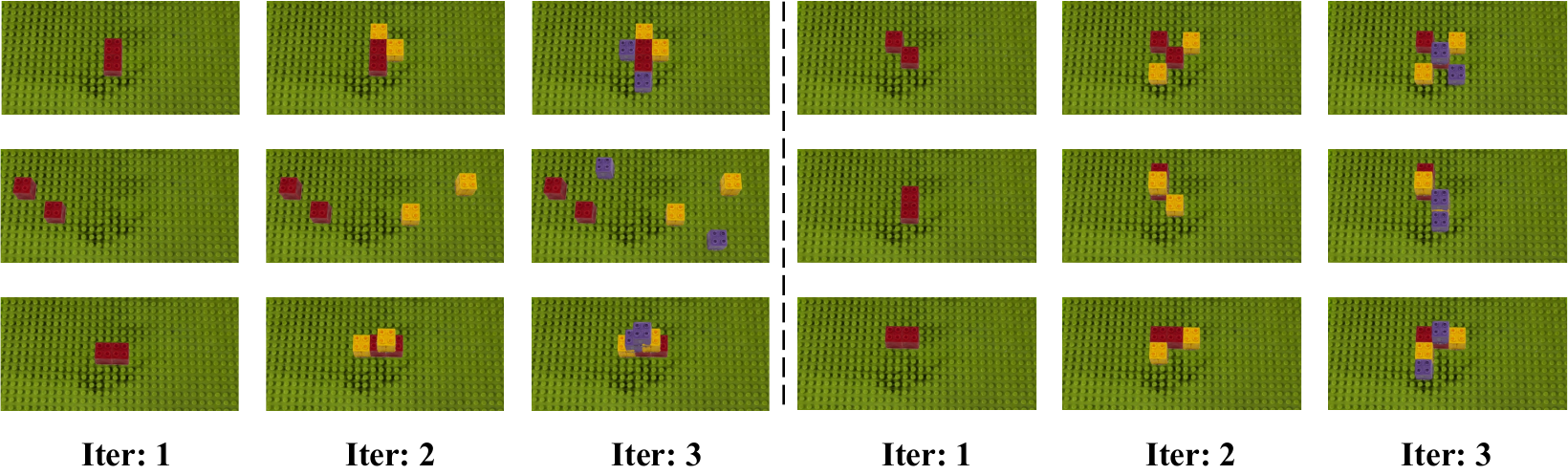}
\caption{\textbf{Iterative manual generation examples for LEGO Assembly.} 
Each row shows a sequence where two bricks are progressively stacked per iteration. 
Scenes are rendered at each step using our digital-twin toolkit.}
\label{fig:appendix_1_more_visualization}
\end{figure*}

\begin{figure*}[t]
\vspace{5pt}
\centering
\includegraphics[width=\textwidth]{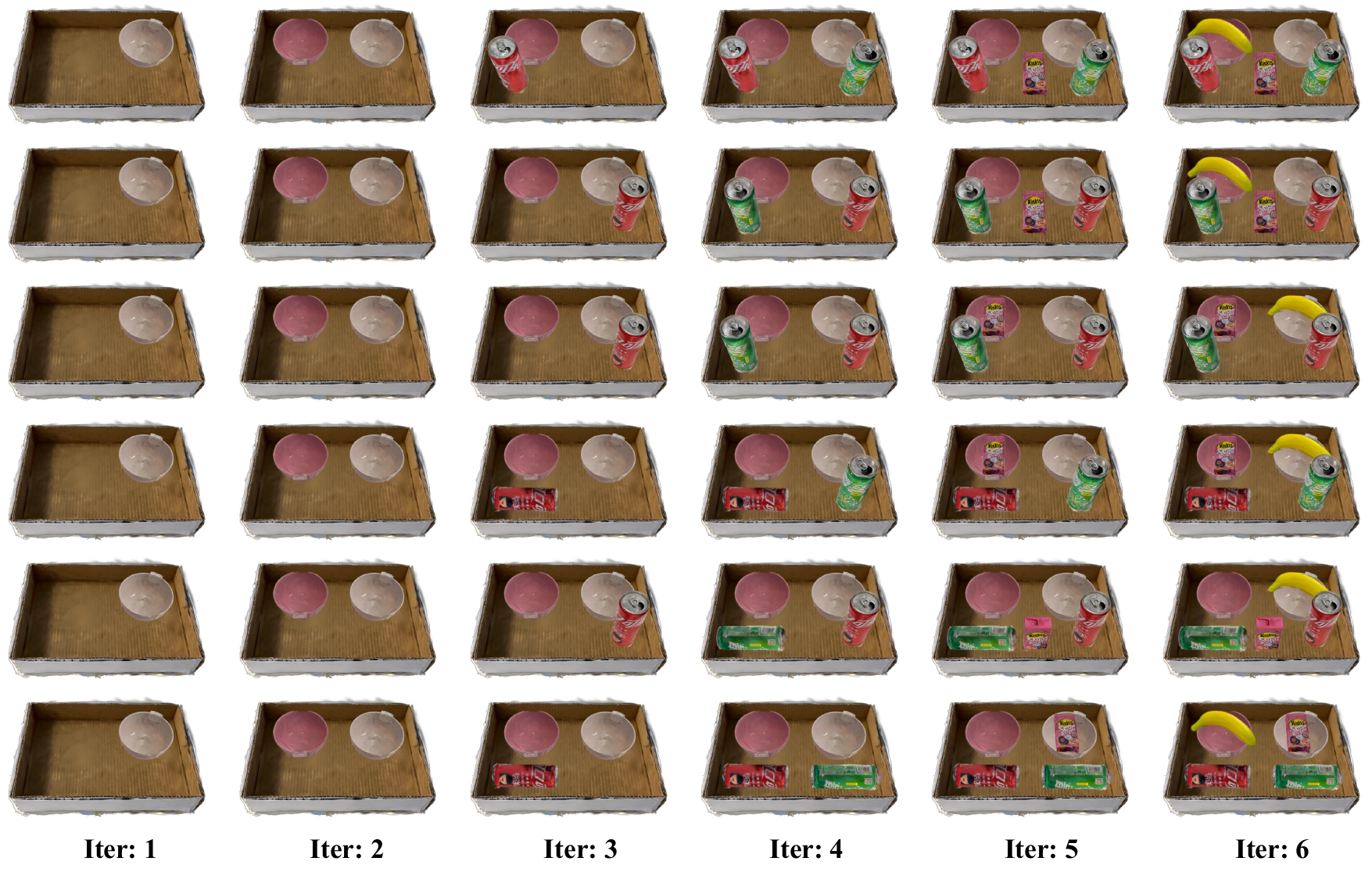}
\caption{\textbf{Iterative manual generation examples for Object Rearrangement.} 
Each row shows a sequential process where objects are placed into the box one at a time.
Scenes are rendered at each step using our digital-twin toolkit.}
\label{fig:appendix_2_more_visualization}
\vspace{5pt}
\end{figure*}

\begin{enumerate}
    \item \textbf{Asset Reconstruction:} We first reconstruct high-fidelity 3D assets of all relevant objects, including the LEGO board, individual bricks of various colors, and the objects used in the rearrangement task. This is achieved by capturing multi-view images of each object and using them to train a 3DGS \cite{2023_8_08-3dgs_for_real_time_radiance_field_rendering} model. The resulting representations are then decomposed and aligned to a unified Cartesian coordinate system for consistent spatial referencing.
    \item \textbf{Iterative Scene Generation:} Given an initial state and a set of available objects, the toolkit iteratively and automatically generates intermediate task states. For LEGO assembly, it sequentially places each brick by randomly sampling a valid position on the board. At each intermediate step, we render a photorealistic image of the current scene from a fixed front-view camera perspective. This process provides the necessary data for training the planning expert: the rendered image serves as the subgoal image, the brick's board position provides the (\textit{U, V}) coordinates, and a corresponding textual description (e.g., "Yellow blocks assemble") is generated via templates.
\end{enumerate}

This automated pipeline enabled us to generate a dataset of over 10,000 frames for each task, providing the rich data needed to effectively pretrain the planning expert. More visualizations of the generated sim-to-real data are shown in Figure \ref{fig:appendix_1_more_visualization} and Figure \ref{fig:appendix_2_more_visualization}.

\section{Additional Experimental Details}
\label{sec:appendix_experiments}

In this section, we report the additional ablation studies on the impact of manual generation quality and token sequence arrangement on model's action accuracy. 

\begin{figure*}[t]
\centering
\vspace{0.4cm}

\begin{minipage}{\textwidth}
  \centering
  
  \includegraphics[width=\textwidth]{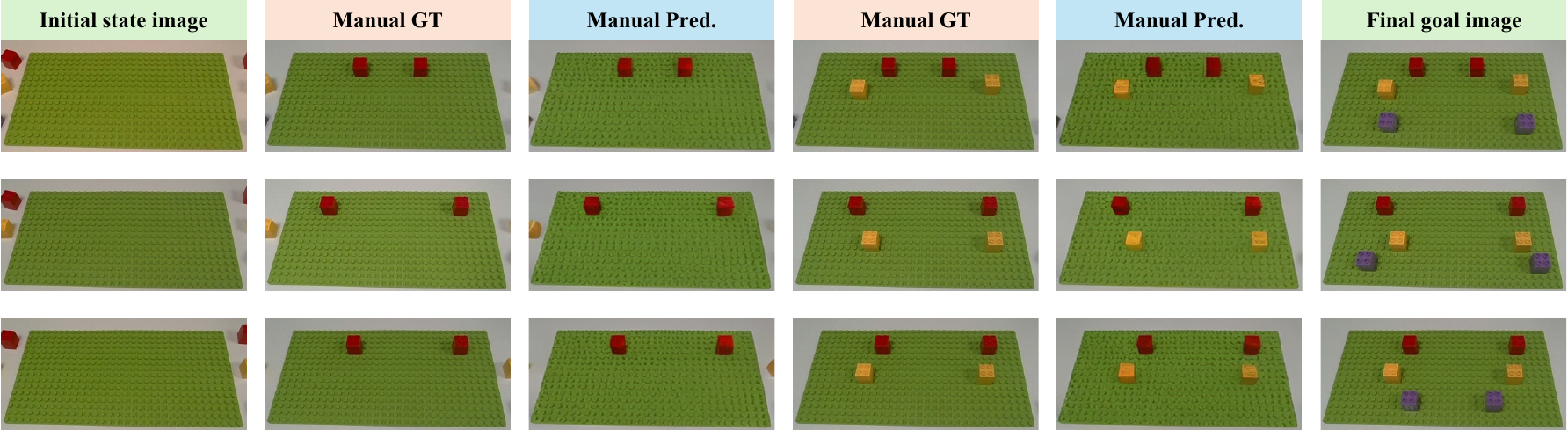}
    \vspace{-0.3cm} 
  \caption{\textbf{Iterative manual generation examples for 2D LEGO Assembly.} Pred refers to the predictions generated by our model, while GT denotes the ground truth in the test set.}
\label{fig:app-gen-vis-1}
\end{minipage}

\vspace{0.3cm}

\begin{minipage}{\textwidth}
  \centering
  \includegraphics[width=\textwidth]{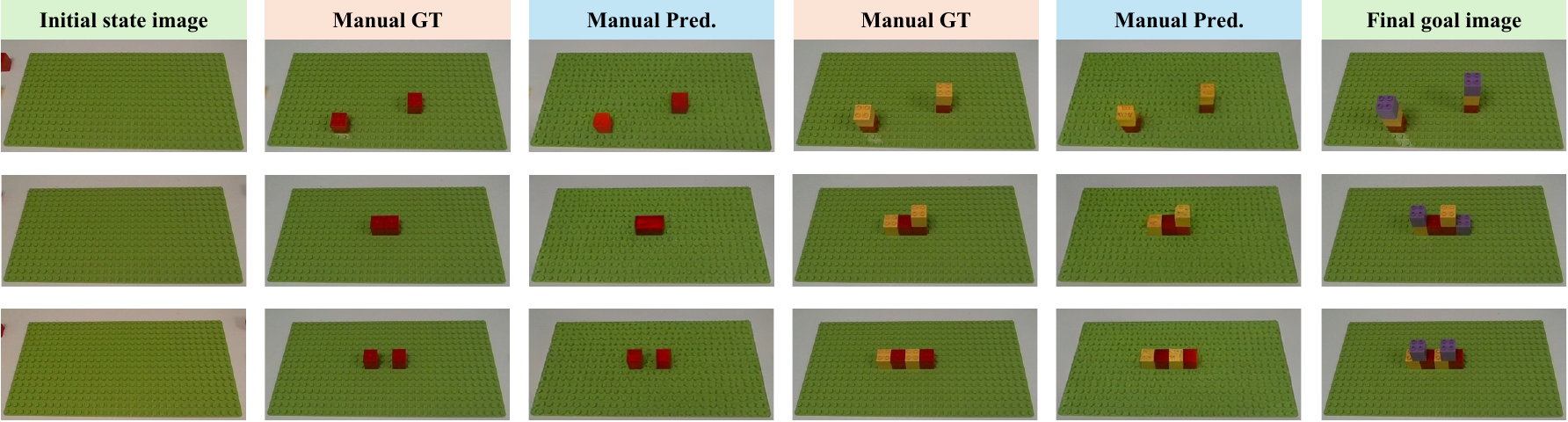}
    \vspace{-0.3cm} 
  
  \caption{\textbf{Iterative manual generation examples for 3D LEGO Assembly.} Pred refers to the predictions generated by our model, while GT denotes the ground truth in the test set.}
\label{fig:app-gen-vis-2}
\end{minipage}

\vspace{0.3cm} 

\begin{minipage}{\textwidth}
  \centering
  \includegraphics[width=\textwidth]{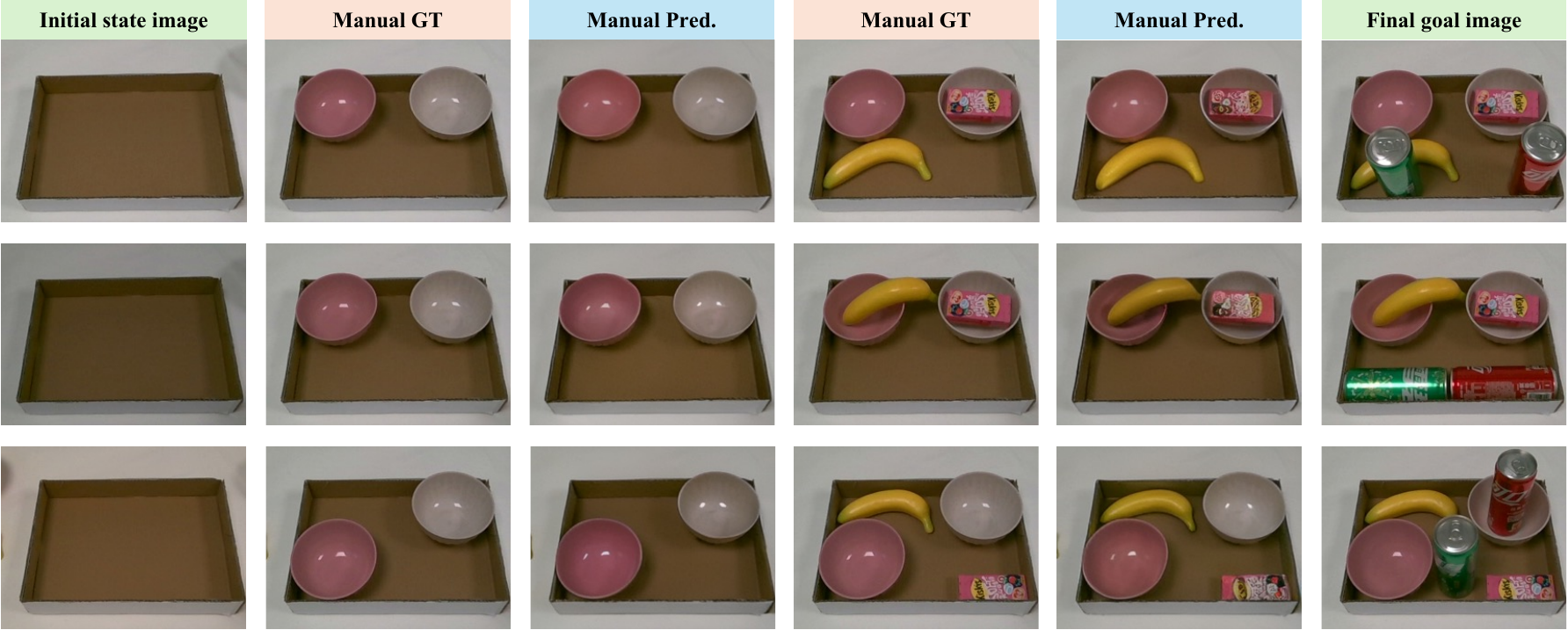}
    \vspace{-0.3cm} 
  
  \caption{\textbf{Iterative manual generation examples for objects rearrangement.} Pred refers to the predictions generated by our model, while GT denotes the ground truth in the test set.}
\label{fig:app-gen-vis-3}
  
\end{minipage}

\end{figure*}
\subsection{Additional Ablation Studies}
This section will present additional ablation studies to further validate our design choices.

\begin{table}[h]
\setlength{\tabcolsep}{1.4pt}
\caption{
\textbf{Comparison of manual generation quality impact on action generation. }
}
\centering
\resizebox{0.48\textwidth}{!}{
\small
\begin{tabular}{c|c|cccc}
\toprule
\multirow{2}{*}{ Training Frames } &
\multirow{2}{*}{ PSNR$\uparrow$ } & 
\multicolumn{4}{c}{\textbf{2D LEGO Assembly}} \\
\cmidrule(lr){3-6}
 &  & 2 bricks $\rightarrow$ & 2 bricks $\rightarrow$ & 2 bricks $\rightarrow$ & \textbf{S.R.} \\
\midrule
0.5K  & 25.71 & 0.35 & 0.25 & 0.20 & 0.20 \\
1K  & 26.61 & 0.45 & 0.35 & 0.30 & 0.25 \\
3K & 27.16 & 0.65 & 0.65 & 0.60 & 0.60 \\
6K & 28.29 & 0.85 & 0.80 & 0.80 & 0.80 \\
\rowcolor[HTML]{FFF0F5}
10K & 29.01 & \textbf{0.95} & \textbf{0.90} & \textbf{0.85} & \textbf{0.85} \\
\bottomrule
\end{tabular}
}
\label{tab:manual_quality}
\end{table}

\subsubsection{Impact of Manual Generation Quality on Action}
To evaluate the robustness of the action expert under varying manual-generation quality, we compare five versions of our planning expert trained on datasets of 0.5K, 1K, 3K, 6K and 10K frames, respectively, for the 2D LEGO Assembly task. All these training frames are generated using our high-fidelity digital-twin toolkit. These planning experts produce manuals with differing levels of fidelity, quantified by the PSNR of the generated subgoal images. We condition ManualVLA on these manuals and measure the resulting task success rate over 20 rollouts, as reported in Table~\ref{tab:manual_quality}. The results indicate that low-quality subgoal images lead to substantial error accumulation during action generation, significantly degrading overall performance. Once the training set reaches 1,000 samples, yielding manuals with PSNR above 27, the action expert exhibits stable and reliable behavior. This trend highlights ManualVLA’s robustness: when the planning expert is sufficiently trained, its explicit and implicit chain-of-thought reasoning reliably supports consistent action generation.

\subsubsection{Impact of Token Sequence Arrangement}

\begin{table}[h]
\setlength{\tabcolsep}{2.0pt}
\caption{
\textbf{Comparison of different token sequence order impact on action generation. }Here, $T$ and $P$ denote the textual description and coordinate pairs $(U, V)$ in the manual, while $I_{\text{subgoal}}$ and $I_{\text{prompt}}$ refer to the subgoal image and the visual prompt image.
}
\centering
\resizebox{0.48\textwidth}{!}{
\begin{tabular}{l|cccc}
\toprule
\multirow{2}{*}{ Token Sequence} &
\multicolumn{4}{c}{\textbf{2D LEGO Assembly}} \\
\cmidrule(lr){2-5}
 & 2 bricks $\rightarrow$ & 2 bricks $\rightarrow$ & 2 bricks $\rightarrow$ & \textbf{S.R.} \\
\midrule
$P$ $\rightarrow$ $I_{subgoal}$ $\rightarrow$ $T$ $\rightarrow$ $I_{prompt}$ & 0.90 & 0.85 & 0.80 & 0.80 \\

$I_{subgoal}$ $\rightarrow$ $P$ $\rightarrow$ $T$ $\rightarrow$ $I_{prompt}$ & 0.75 & 0.75 & 0.70 & 0.70 \\


$T$ $\rightarrow$ $I_{subgoal}$ $\rightarrow$ $P$ $\rightarrow$ $I_{prompt}$ & 0.85 & 0.80 & 0.80 & 0.80 \\

\rowcolor[HTML]{FFF0F5}
$T$ $\rightarrow$ $P$ $\rightarrow$ $I_{subgoal}$  $\rightarrow$ $I_{prompt}$  & \textbf{0.95} & \textbf{0.90} & \textbf{0.85} & \textbf{0.85} \\
\bottomrule
\end{tabular}
}
\label{tab:token_sequence}
\end{table}

To further analyze how the ordering of multimodal tokens in the generated manual influences action generation, we evaluate four different sequence arrangements, as shown in Table~\ref{tab:token_sequence}. The modalities contained in the manual serve complementary purposes: the text instruction provides high-level semantic goals, the coordinate tokens $(U, V)$ specify the future spatial locations of the objects to be manipulated, and the generated manual image offers step-wise visual cues synthesized by the planning expert. Finally, the visual prompt image is conditioned on the generated coordinates $(U, V)$, and therefore we place it at the end of the token sequence. This ordering naturally forms a pipeline in which the model first performs implicit CoT reasoning, followed by explicit CoT reasoning, before producing the final action sequence.
Because the effectiveness of the action expert depends on how well it integrates semantic and visual information, the ordering of these tokens can substantially affect downstream policy performance. Our study on different generation orders of the three types of manual information reveals that the sequence of generating text first, then coordinates, and finally subgoal images yields the best task success rate. Meanwhile, the other sequence configurations introduce only minor performance degradation, demonstrating both the robustness of ManualVLA and the critical role of the combined implicit–explicit CoT reasoning process in enabling strong action-generation performance.

\begin{figure*}[t]
    \centering
    \includegraphics[width=0.98\textwidth]{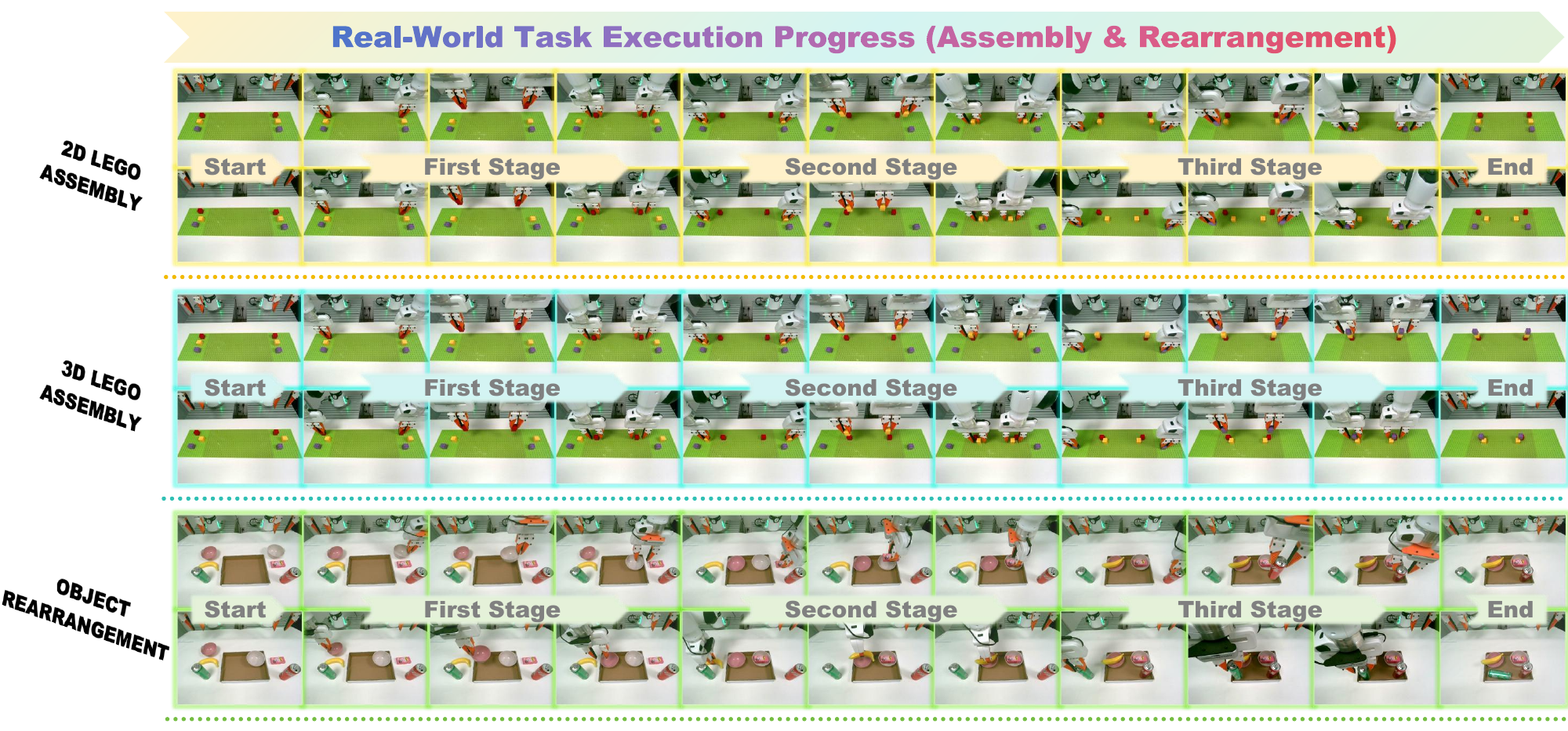} 
    \vspace{-0.3cm}
    \caption{\textbf{Real-World Task Execution Progress Visualization.} We provide visualizations of three real world tasks including assembly and rearrangement evaluated on dual-arm Franka robot platform. }
    \label{fig:execution}
    \vspace{-0.3cm}
\end{figure*}

\begin{figure*}[t]
\centering
\vspace{-0.2cm}
\includegraphics[width=0.95\textwidth]{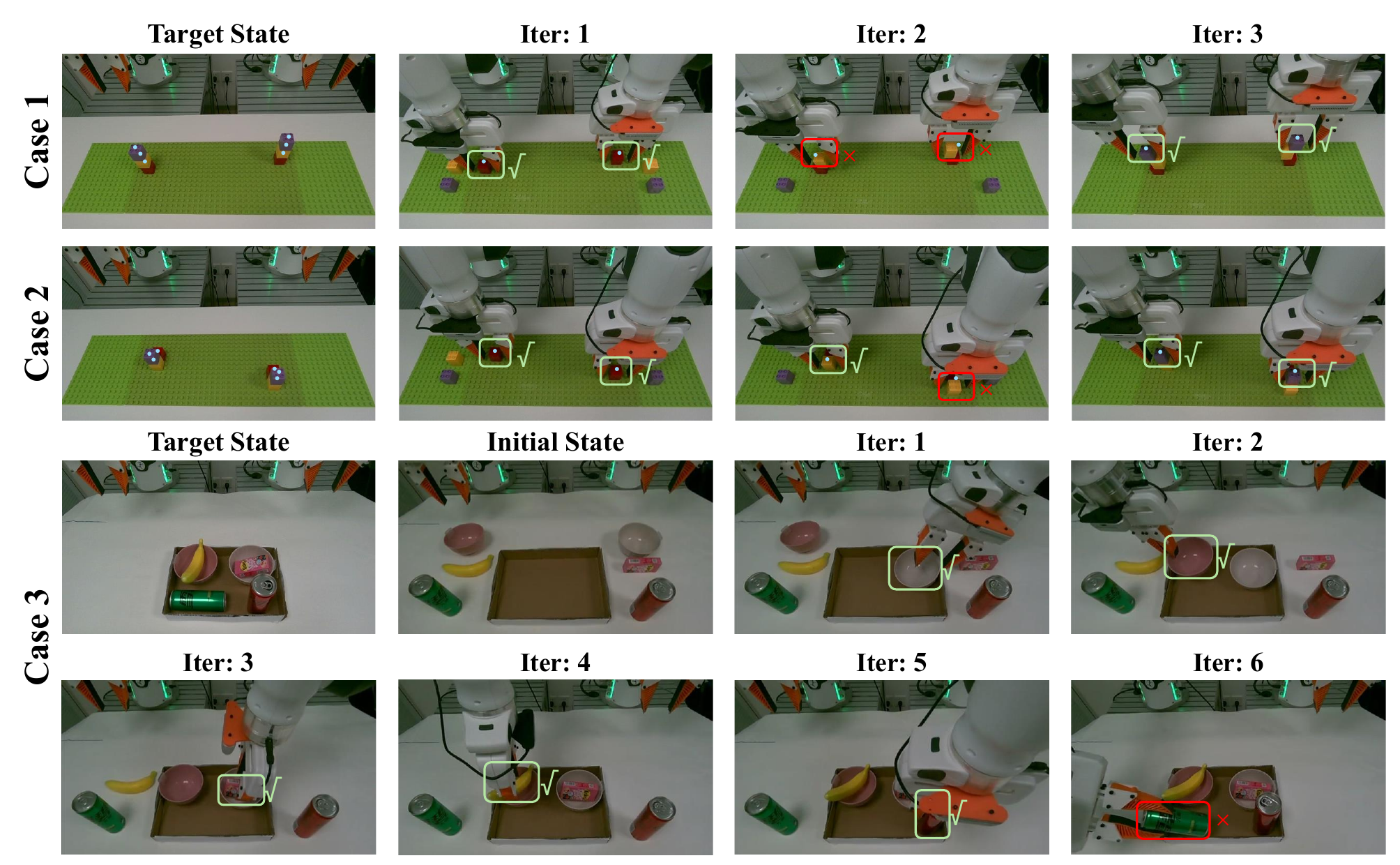}
\vspace{-0.3cm}
\caption{\textbf{Failure cases in our two tasks: LEGO assembly and objects rearrangement.} 
The top two rows illustrates two LEGO failure cases and the bottom two rows shows a failure case of objects rearrangement task.}
\vspace{-0.3cm}
\label{fig:appendix_3_failing_cases}
\end{figure*}

\section{Additional Qualitative Results}
\label{sec:appendix_visualizations}

This section provides more qualitative results for manual generation and real-world robot rollouts.

\subsection{Manual Generation Visualization}
Figure~\ref{fig:app-gen-vis-1}, Figure~\ref{fig:app-gen-vis-2}, and Figure~\ref{fig:app-gen-vis-3} provide additional visualizations of the manuals generated by our planning expert across all three downstream tasks. These examples showcase the model's ability to generate structured and interpretable intermediate states that accurately guide the subsequent action generation. For the LEGO assembly tasks, ManualVLA sequentially reconstructs the correct brick placements and colors, demonstrating precise step-wise reasoning. Similarly, for object rearrangement, it progressively generates subgoals that accurately capture the spatial relationships between objects, moving step-by-step toward the final goal configuration. Overall, these results highlight ManualVLA's strong intermediate reasoning capabilities, establishing a reliable foundation for the action expert to generate accurate actions.

\subsection{Real-World Rollout Visualization}
The qualitative rollouts in Figure~\ref{fig:execution} further corroborate our quantitative findings, illustrating keyframes of the dual-arm real-world execution processes. The visualizations demonstrate that ManualVLA can follow the internally generated manuals to reliably guide the action expert in producing precise grasping, insertion, and placement motions. 
In both the 2D and 3D LEGO assembly tasks, compared with the final goal image, the robot consistently maintains accurate brick placement throughout all stages. For the object-rearrangement task, also compared with the final goal image, it stably manipulates objects with varying shapes and occlusions.
These results collectively validate ManualVLA’s strong action generation capabilities, demonstrating its potential as a robust policy for real-world, long-horizon robotic manipulation.

\section{Failure Case Analysis}
\label{sec:appendix_failure}
Although ManualVLA demonstrates strong overall performance, it is not without limitations. Through our experiments, we identified two primary failure modes, as illustrated in Figure~\ref{fig:appendix_3_failing_cases}:

\begin{enumerate}
    \item \textbf{Occasional Erroneous LEGO Placement.} While ManualVLA is generally successful in accomplishing the LEGO assembly task, it can still produce incorrect placements. As shown in Cases 1 and 2 of Figure~\ref{fig:appendix_3_failing_cases}, the system places the yellow bricks in incorrect positions due to model errors. Notably, however, the system is often able to recover from such mistakes and correctly place subsequent bricks.

    \item \textbf{Placement Errors Under Large Rotation Angles.} In the Objects Rearrangement task, certain scenarios require the robot arm to perform large rotations to achieve the correct placement orientation. ManualVLA may fail in these situations, as illustrated by Case 3 in Figure~\ref{fig:appendix_3_failing_cases}, where the robot arm fails to place the spirit can into the box. We hypothesize that these failures stem from the limited number of such extreme rotation cases in the training data.
\end{enumerate}

\end{document}